\documentclass[lettersize,journal]{IEEEtran}
\usepackage{amsmath,amsfonts}
\usepackage{algorithmic}
\usepackage{algorithm}
\usepackage{array}
\usepackage[caption=false,font=normalsize,labelfont=sf,textfont=sf]{subfig}
\usepackage{textcomp}
\usepackage{stfloats}
\usepackage{url}
\usepackage{verbatim}
\usepackage{graphicx}
\usepackage{cite}
\usepackage{bbm}
\usepackage{multirow}
\usepackage{booktabs}
\usepackage{xcolor}
\usepackage{makecell}
\usepackage{bbding}
\usepackage{colortbl}
\usepackage{caption}
\usepackage{afterpage}
\usepackage{capt-of}
\usepackage{tcolorbox}

\definecolor{light-red}{rgb}{1.0, 0.7, 0.7}
\definecolor{light-yellow}{rgb}{1.0, 1.0, 0.8}
\definecolor{light-orange}{rgb}{1.0, 0.9, 0.8}
\definecolor{scannet-color}{rgb}{0.945, 0.894, 0.824}
\definecolor{replica-color}{rgb}{0.827, 0.945, 0.957}
\definecolor{hypersim-color}{rgb}{0.835, 0.796, 0.929}
\definecolor{things-color}{rgb}{0.843, 0.941, 0.941}

\begin{document}

\title{PLGS: Robust Panoptic Lifting with 3D Gaussian Splatting}

\author{Yu Wang,
        Xiaobao Wei,
        Ming Lu,
        Guoliang Kang
\thanks{Yu Wang, School of Automation Science and Electrical Engineering, Beihang University, Beijing, 100191, China, e-mail: wangyuyy@buaa.edu.cn.

Xiaobao Wei, Institute of Software, Chinese Academy of Sciences \& University of Chinese Academy of Sciences, e-mail: weixiaobao0210@gmail.com.

Ming Lu, Peking University, e-mail: lu199192@gmail.com.

Guoliang Kang, School of Automation Science and Electrical Engineering, Beihang University, e-mail: kgl.prml@gmail.com.

(Corresponding author: Guoliang Kang)

}

\thanks{}}


\maketitle

\begin{abstract}
Previous methods utilize the Neural Radiance Field (NeRF) for panoptic lifting, while their training and rendering speed are unsatisfactory. In contrast, 3D Gaussian Splatting (3DGS) has emerged as a prominent technique due to its rapid training and rendering speed. However, unlike NeRF, the conventional 3DGS may not satisfy the basic smoothness assumption as it does not rely on any parameterized structures to render (e.g., MLPs). Consequently, the conventional 3DGS is, in nature, more susceptible to noisy 2D mask supervision. In this paper, we propose a new method called PLGS that enables 3DGS to generate consistent panoptic segmentation masks from noisy 2D segmentation masks while maintaining superior efficiency compared to NeRF-based methods. Specifically, we build a panoptic-aware structured 3D Gaussian model to introduce smoothness and design effective noise reduction strategies. For the semantic field, instead of initialization with structure from motion, we construct reliable semantic anchor points to initialize the 3D Gaussians. We then use these anchor points as smooth regularization during training. Additionally, we present a self-training approach using pseudo labels generated by merging the rendered masks with the noisy masks to enhance the robustness of PLGS. For the instance field, we project the 2D instance masks into 3D space and match them with oriented bounding boxes to generate cross-view consistent instance masks for supervision. Experiments on various benchmarks demonstrate that our method outperforms previous state-of-the-art methods in terms of both segmentation quality and speed. 
\end{abstract}


\section{Introduction}
\begin{figure}[!t]
    \centering
    \includegraphics[width=0.85\linewidth]{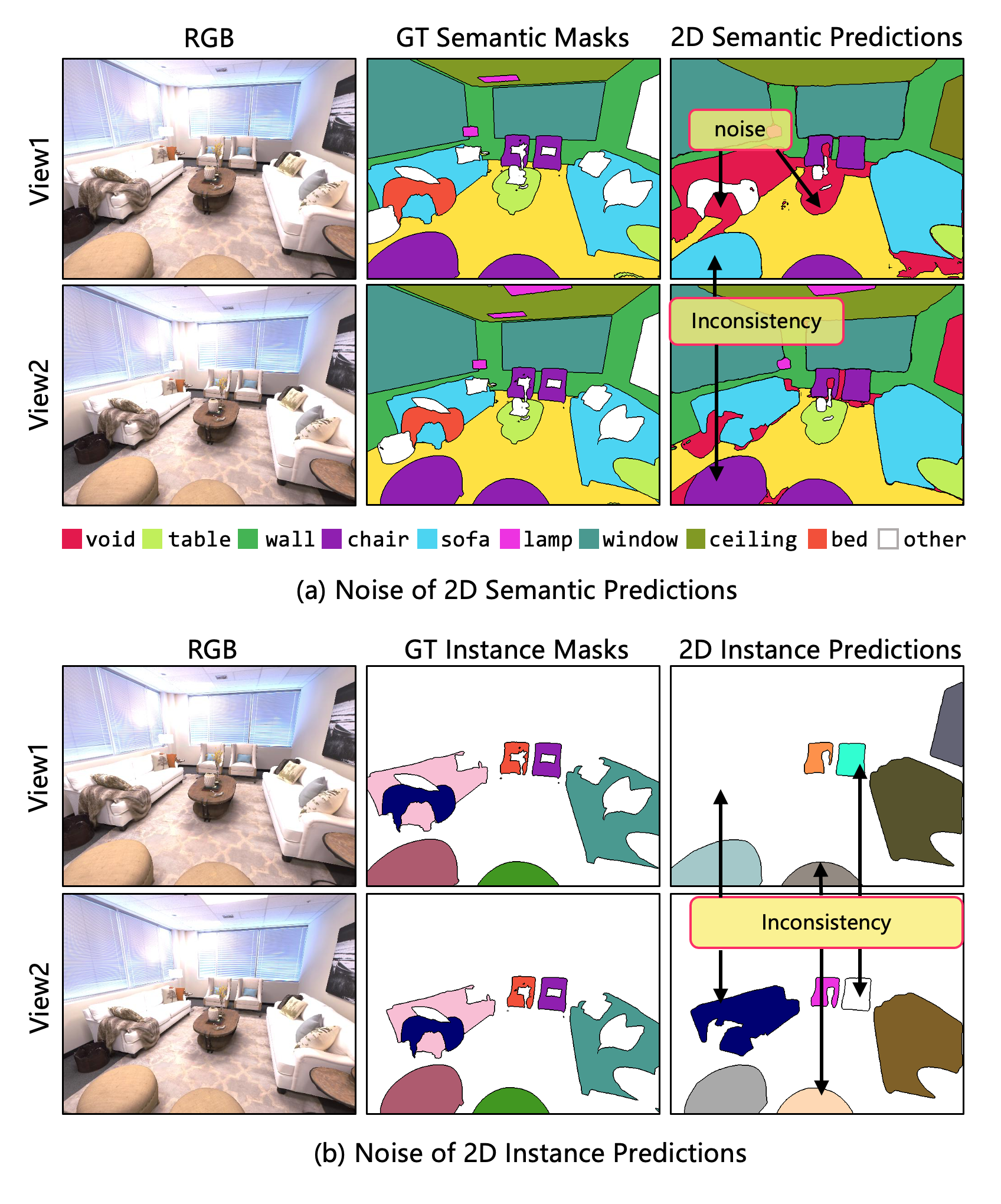}
    \caption{Noise of 2D-generated masks with Mask2Former~\cite{cheng2022masked}. 
    There is noticeable noise even with a slight change of viewpoint.
    (a) For semantic masks, there are invalid predictions (\emph{e.g.} sofa and table) and inconsistent predictions (\emph{e.g.} chair misclassified to sofa).
    (b) For instance masks, the instance labels of the masks vary inconsistently across different views.
    }
    \label{fig:noisy-pattern}
\end{figure}

    \IEEEPARstart{R}{obust} 3D panoptic scene understanding plays an essential role in various applications, \emph{e.g.}, robot grasping, self-driving, \emph{etc.} 
    Though rapid progress has been made in 2D panoptic segmentation tasks, it is still challenging to obtain 3D panoptic segmentation masks of a specific scene which should be consistent in both semantic-level and instance-level across different views. 
    The reasons are two-fold. 
    Firstly, it is hard to directly apply the 2D panoptic segmentation model to images of different views to achieve consistent segmentation masks. 
    It is because single-image segmentation model does not have a basic understanding of the 3D scene and is sensitive to the view variations, thus generating noisy and view-inconsistent masks, as shown in Fig.~\ref{fig:noisy-pattern}.
    Secondly, to gain a full understanding of the 3D scene, we need large amounts of 2D or 3D annotations~\cite{zhi2021place, wang2022dm, he2023prototype, tao2022seggroup, sun2024image, shuai2021backward, gasperini2021panoster, dahnert2021panoptic, milioto2020lidar}, which are expensive and time-consuming to obtain. 
    Therefore, a typical way to perform 3D panoptic segmentation is to lift the segmentation masks from 2D to 3D~\cite{siddiqui2023panoptic, bhalgat2023contrastive}, \emph{i.e.,} training a 3D panoptic segmentation model based on machine-generated noisy and view-inconsistent 2D segmentation masks across different views.

\begin{figure*}[!t]
    \includegraphics[width=\textwidth]{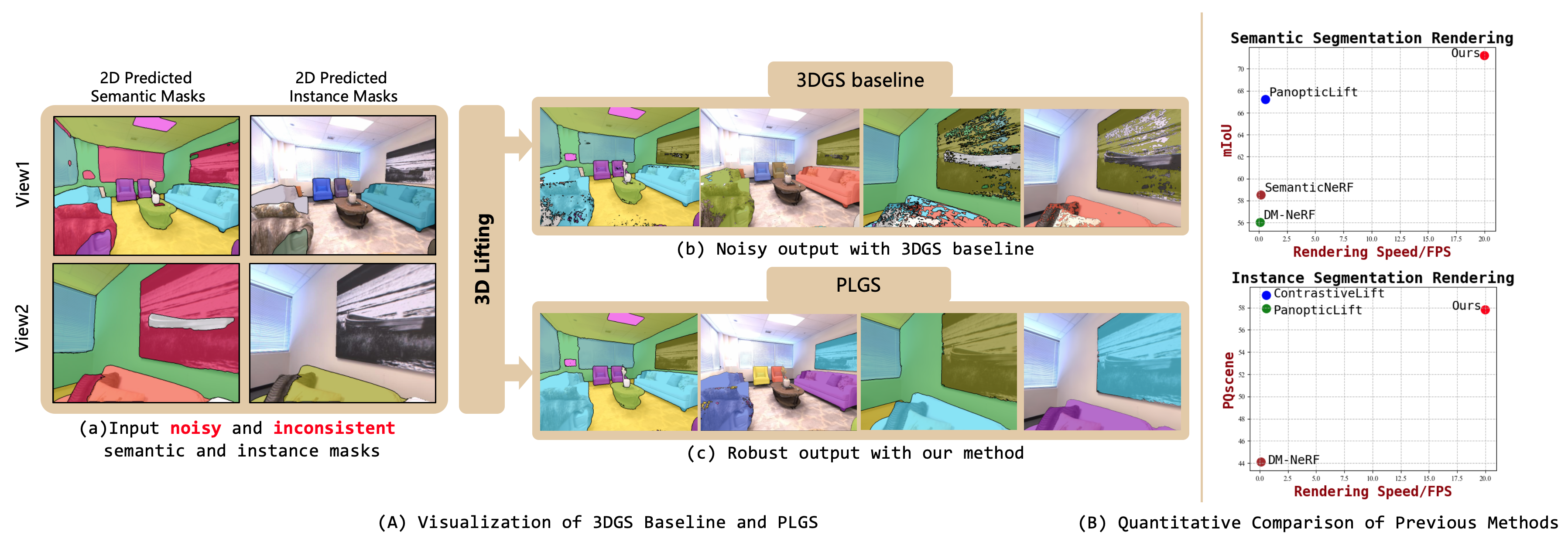}
    \caption{Given noisy and inconsistent machine-generated panoptic masks, PLGS is capable of robustly and rapidly lifting 2D masks into 3D panoptic field, achieving consistent 3D panoptic segmentation with high rendering speed. (A) demonstrates that PLGS achieves robust results of panoptic field reconstruction with much less noise compared to 3DGS baseline. (B) exhibits that PLGS achieves comparable panoptic reconstruction quality with much faster rendering speed compared to previous state-of-the-art methods.}
    \label{fig:introduction}
\end{figure*}

    Specifically, Panoptic Lifting~\cite{siddiqui2023panoptic} and Contrastive Lift~\cite{bhalgat2023contrastive} utilize Neural Radiance Field (NeRF)~\cite{mildenhall2021nerf} and its derived model~\cite{chen2022tensorf} to lift machine-generated segmentation masks from 2D to 3D, enabling photo-realistic rendering of consistent panoptic segmentation masks across novel views. 
    Typically, given a set of images of a scene, they use powerful 2D panoptic segmentation model, \emph{i.e.} Mask2Former~\cite{cheng2022masked}, to generate semantic and instance masks which are noisy and inconsistent as shown in Fig.~\ref{fig:noisy-pattern}.
    To deal with the noisy supervision, Panoptic Lifting~\cite{siddiqui2023panoptic} enhances the reliability of the input masks with a Test-Time Augmentation (TTA) strategy and utilizes Hungarian Algorithm~\cite{kuhn1955hungarian} for instance supervision.
    Building on this, Contrastive Lift~\cite{bhalgat2023contrastive} introduces contrastive loss and mean teacher strategy, significantly improving instance segmentation quality.
    Despite promising progress achieved, limited to the pattern of NeRF, the training and rendering speed is still quite low and far from satisfactory.

    Recently, 3D Gaussian Splatting (3DGS)~\cite{kerbl20233d} has demonstrated superior rendering quality and speed compared to NeRF. 
    For example, 3DGS outperforms instant-NGP~\cite{muller2022instant} in terms of rendering FPS in over 150 FPS. 
    Therefore, it is natural to replace NeRF with 3DGS in typical NeRF-based panoptic lifting frameworks to improve training and rendering speeds.
    However, as shown in Fig.~\ref{fig:introduction}(A), while 3DGS significantly accelerates these processes, it produces inferior segmentation masks compared to NeRF-based methods, with noticeable noise and inconsistencies of semantic and instance masks across different views.
    Moreover, naively transferring previous noise handling techniques (\emph{e.g.,} using test-time augmented data and segment consistency loss) in NeRF-based panoptic lifting methods to 3DGS-based panoptic lifting cannot obtain satisfactory results (see Sec.~\ref{sec:quality comparison}).
    It may be because 3DGS does not rely on any parameterized structures (\emph{e.g.,} MLPs adopted in typical NeRF) and may break the basic smoothness assumption in terms of the mapping from features to masks.

    In this paper, we introduce PLGS, a novel method that robustly lifts noisy panoptic masks of in-place scenes with rapid speed with the help of accessible depth maps.
    Firstly, we build a structured panoptic-aware Gaussian model based on Scaffold-GS~\cite{lu2023scaffold} instead of vanilla 3DGS to introduce smoothness.
    Furthermore, we design effective strategies to ensure consistency in both semantic and instance mask predictions across views. 
    For semantic field reconstruction, we initialize the model with a robust semantic point cloud, derived through a voting mechanism from machine-generated masks, rather than the sparse RGB point cloud from COLMAP~\cite{schonberger2016structure}. 
    This semantic point cloud, further refined with locally consistent 2D masks in datasets with continuous viewpoints, initializes PLGS through voxelization and regularization.
    Additionally, to enhance the performance of PLGS, we introduce an effective self-training strategy using pseudo labels as supervision. 
    The pseudo labels are generated by integrating the segments of machine-generated masks and rendered masks with consistent semantic labels.
    For instance field reconstruction, we introduce an effective method to match inconsistent instance masks in 3D space, ensuring robust and uniform instance masks.
    
    Our main contributions can be summarized as follows:
    \begin{itemize}

    \item We introduce a new framework named PLGS to lift noisy masks from 2D to 3D using 3DGS, enabling the rapid generation of consistent panoptic segmentation masks across views without ground-truth annotations. 
    
    \item To mitigate the negative effects of noisy and inconsistent mask supervision, we structure the 3D Gaussian to introduce smoothness and design effective noise reduction strategies for both semantic and instance segmentation, enhancing the robustness of our approach.

    \item Extensive experiments on HyperSim~\cite{roberts2021hypersim}, Replica~\cite{straub2019replica} and ScanNet~\cite{dai2017scannet} datasets demonstrate that PLGS outperforms previous SOTA methods in terms of panoptic segmentation quality and training/rendering speed.
    
    \end{itemize}

\section{Related Work}
\noindent \textbf{Neural Scene Representation.}
    Recent years have witnessed significant advancements in neural scene representation techniques. NeRF~\cite{mildenhall2021nerf}, for instance, leverages neural networks to encode the geometry and appearance of 3D scenes and employs differentiable ray-marching volume rendering for training, allowing for photo-realistic novel view synthesis with only posed 2D images as input.
    Subsequent research efforts have introduced various enhancements, such as NeRF++~\cite{zhang2020nerf++}, MipNeRF~\cite{barron2021mip}, MipNeRF360~\cite{barron2022mip}, and Tri-MipRF~\cite{hu2023tri}, which aim to improve rendering quality. 
    On the other hand, approaches like Plenoxel~\cite{fridovich2022plenoxels}, DVGO~\cite{sun2022direct}, TensoRF~\cite{chen2022tensorf}, and instant-NGP~\cite{muller2022instant} focus on accelerating the optimization process for neural representations.
    Recently, 3D Gaussian Splatting (3DGS)~\cite{kerbl20233d} has emerged as a promising approach for scene representation, utilizing 3D Gaussians to achieve high-quality and real-time rendering. 
    Based on 3DGS, Scaffold-GS~\cite{lu2023scaffold} proposes to enhance the robustness to view changes by introducing a compact structure to distribute local 3D Gaussians.
    Therefore, we build our PLGS based on Scaffold-GS to take both robustness and speed into account.

\noindent \textbf{Semantic Aware Neural Scene Representation.}
    Several approaches have been developed to extend NeRF for 3D scene understanding, particularly in the segmentation domain. 
    Semantic-NeRF~\cite{zhi2021place}, PNF~\cite{kundu2022panoptic} and DM-NeRF~\cite{wang2022dm} enable the concurrent optimization of radiance and segmentation fields to achieve 3D-consistent scene segmentation, which rely on the costly dense-view annotated labels. 
    While several works~\cite{kerr2023lerf, zhou2024feature, qin2024langsplat, liu2023weakly, shi2024language} lift latent features of 2D foundation model~\cite{radford2021learning} into 3D for open-vocabulary queries, they primarily focus on semantic segmentation, neglecting instance-aware modeling and the cross-view inconsistency issue.
    Some other works~\cite{ye2023gaussian,shen2023distilled,goel2023interactive,wei2023noc,cen2023segment,kim2024garfield} utilize 2D object-level segmentation models~\cite{kirillov2023segment} to achieve object-segmentation, which though instance-aware, lacks holistic understanding of the scene and the corresponding semantic label of each object.

    Different from the above methods, panoptic lifting with 2D machine-generated panoptic masks is not only semantic- and instance-aware but also practical in common use. 
    However, it faces challenges of dealing with noisy semantic labels and inconsistent instance IDs across viewpoints.
    Previous work Panoptic Lifting~\cite{siddiqui2023panoptic}, built on TensoRF~\cite{chen2022tensorf} effectively reduces the inconsistency by introducing methods like TTA pre-processing strategy, segment consistency loss for semantic segmentation and linear assignment strategy for instance segmentation.
    Based on Panoptic Lifting, instead of utilizing linear assignment strategy, Contrastive Lift~\cite{bhalgat2023contrastive} introduces contrastive loss and mean-teacher training strategy to achieve superior performance of instance segmentation.
    However, limited by the rendering mechanism of NeRF, both methods suffer from huge training time consumption and low rendering speed.
    Therefore, we propose to utilize 3DGS to achieve robust and high-quality panoptic lifting with much less training time required and faster rendering speed.

\noindent \textbf{2D and 3D Panoptic Segmentation.}
    In recent years, many works~\cite{cheng2022masked, kirillov2019panoptic, zhang2021k, yang2019deeperlab, xiong2019upsnet, li2023mask, cheng2020panoptic, wang2020axial, chen2022vision} have been proposed for 2D panoptic segmentation. 
    Typically, most methods use the transformer-based network trained with large scale datasets (\emph{e.g.} COCO~\cite{lin2014microsoft}, ADE20K~\cite{zhou2017scene}, KITTI~\cite{geiger2012we}) to predict semantic mask and instance mask for each input image.
    However, these 2D methods focus on single image segmentation and lack generalization ability of viewpoint variations, resulting in inconsistent segmentation masks with even slight change of viewpoint, which are therefore insufficient for 3D scene panoptic segmentation.

    There are also several works~\cite{he2023prototype, tao2022seggroup, sun2024image, shuai2021backward, gasperini2021panoster, dahnert2021panoptic, milioto2020lidar, narita2019panopticfusion, zhou2021panoptic, wu2024point, wang2024panoptic, zhou2021panoptic} proposed for 3D panoptic segmentation.
    However, these methods take specific 3D context (\emph{e.g.}point cloud, voxel grid, mesh) as input, which makes them incapable of generating dense scene-level panoptic masks.
    Moreover, the 3D panoptic segmentation models demand large scale 3D annotation for training, which are much more expensive than 2D annotations, resulting in inferior generalization ability compared to 2D models.
    Therefore, we propose to effectively lift the panoptic masks generated by the off-the-shelf 2D panoptic segmentation models with no annotation needed.

\begin{figure*}[htb]
    \centering
    \includegraphics[width=0.85\linewidth]{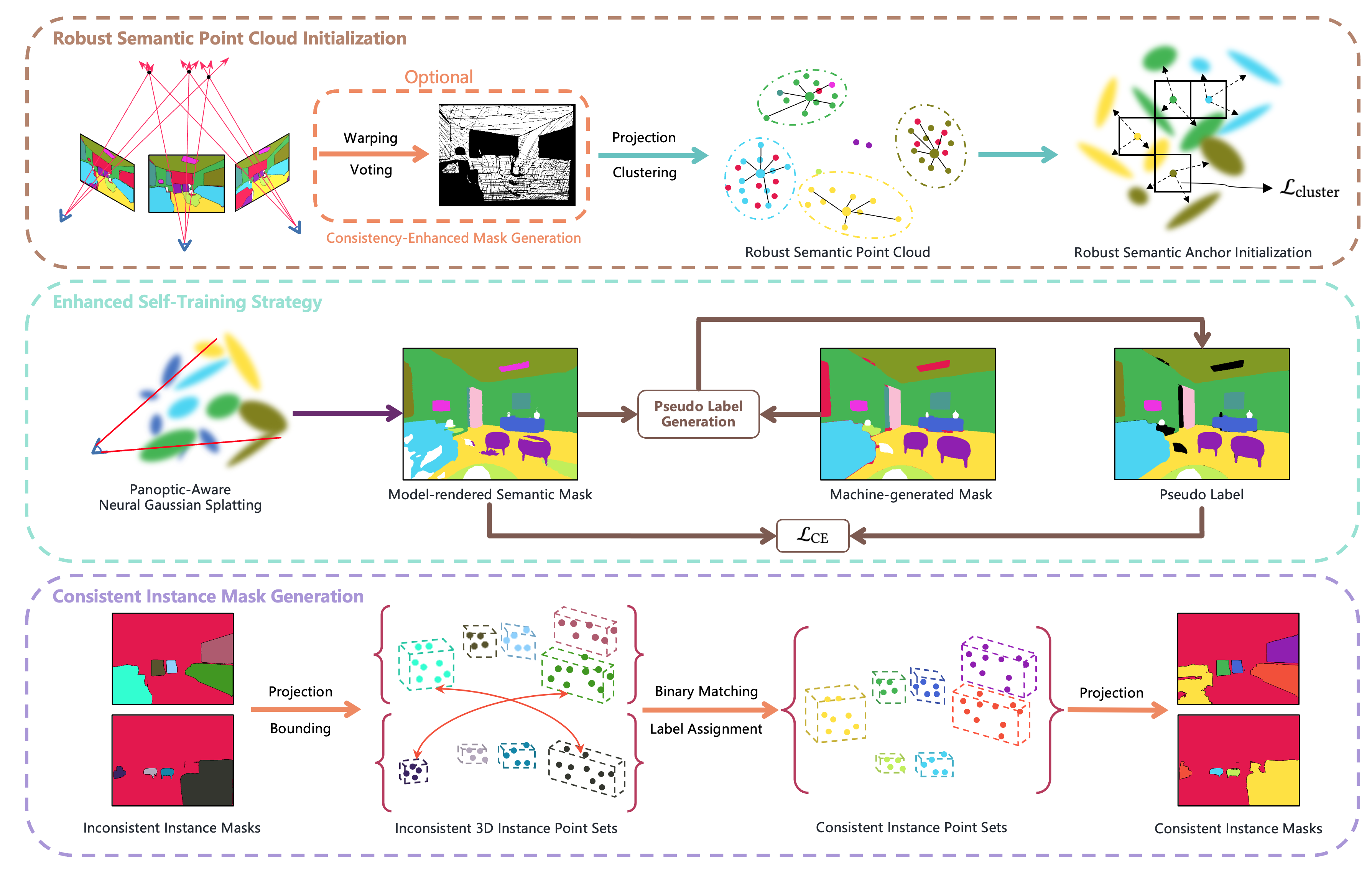}
    \caption{The general framework of PLGS. We build our model based on Scaffold-GS~\cite{lu2023scaffold} with additional semantic and instance decoder. 
    For semantic reconstruction, we project and cluster reliable semantic labels of the machine-generated masks in 3D space to generate robust semantic point cloud for initialization. 
    We further design an enhanced self-training strategy to integrate the segments of the input semantic masks and the consistent semantic labels of the model-rendered semantic masks to promote the performance of the reconstruction. 
    For instance reconstruction, we propose to match the inconsistent instance masks in 3D space to generate consistent instant masks for supervision.}
    \label{fig:pipeline}
\end{figure*}

\section{Method}

    Given a set of RGB images $\{\text{I}_i\}_{i=1}^N$ and depth maps $\{\text{D}_i\}_{i=1}^N$ with camera intrinsics and extrinsics of a static scene, PLGS aims to lift noisy and view-inconsistent machine-generated panoptic masks into a robust 3D panoptic representation with rapid training and rendering speed. 
    We build the panoptic-aware PLGS based on structured Scaffold-GS~\cite{lu2023scaffold} to introduce smoothness.
    Instead of initializing PLGS with a sparse point cloud generated by COLMAP~\cite{schonberger2016structure}, we incorporate a reliable semantic point cloud generated by projecting and clustering semantic labels in 3D space as shown in Fig.~\ref{fig:pipeline}. 
    We further enhance the semantic reconstruction quality of PLGS by introducing a self-training strategy that integrates input segments with consistently rendered semantic masks. 
    For instance segmentation, we propose to match the inconsistent instance masks in 3D space with oriented bounding boxes to generate consistent instance masks for supervision.

\subsection{Preliminary}\label{sec:preliminary}
    3D Gaussian Splatting~\cite{kerbl20233d} is an explicit model that represents a 3D scene with a set of 3D Gaussians along with a differentiable rasterization technique, demonstrating outstanding reconstruction quality and real-time performance. 
    Each 3D Gaussian is equipped with learnable parameters including center positions, scaling matrix, rotation matrix, opacity and view-depending color. 
    The scaling matrix $\mathbf{S}$ and rotation matrix $\mathbf{R}$ describe the covariance matrix $\mathbf{\Sigma} = \mathbf{R}\mathbf{S}\mathbf{S}^T\mathbf{R}^T$.
    During training, the 3D Gaussians are splatted onto 2D image plane following the instruction in~\cite{zwicker2001ewa}.
    Then the 2D images are rendered by $\alpha$-compositing:
    \begin{equation}\label{eq:render equation}
        C = \sum_{i\in\mathcal{N}}c_i \alpha_i'\prod_{j=1}^{i-1}(1-\alpha_j')
    \end{equation}
    where $C$ represents the color of an arbitrary pixel, $\mathcal{N}$ denotes the set of Gaussians covering the pixel after splatting, $c_i$ denotes the color of the Gaussians, $\alpha_i'$ denotes the effective opacity of the $i$-th projected Gaussian which is obtained by multiplying $\alpha_i$ and the 2D Gaussian distribution. The model is then trained with L1 loss and SSIM~\cite{wang2004image} loss.

\subsection{Structured Panoptic-Aware Scene Representation}\label{sec:structured scene representation} 
    Although 3DGS has achieved splendid reconstruction quality with rapid training and rendering speed, it's infeasible to solve the problem with noisy input due to its non-parameterized structures.
    Therefore, in order to introduce smoothness, we build a structured panoptic-aware model based on Scaffold-GS~\cite{lu2023scaffold} which incorporates a set of anchor points to determine 3D Gaussians.
    We initialize the model with a robust semantic point cloud as described in Sec.~\ref{sec:robust semantic initialization}.
    Given an initial point cloud $\textbf{P} \in \mathbb{R}^{N_\text{pcd} \times 3}$, we generate the anchor points $\textbf{V}\in \mathbb{R}^{N_\text{ach}\times3}$ by voxelization:
    \begin{equation}
        \textbf{V}=\left\{\left[\frac{\textbf{P}}{\epsilon}\right]\right\}\cdot \epsilon
    \end{equation}
    where $\epsilon$ denotes the voxel size, $\left[ \cdot \right]$ denotes the rounding operation and $\left\{\cdot\right\}$ denotes the deduplication operation, $N_\text{pcd}$ and $N_\text{ach}$ denotes the number of the point cloud and the anchor points respectively. 
    Each anchor point $\mathbf{v}_i\in \mathbb{R}^{1\times3}, i\in \{1,2, \cdots, N_{\text{ach}}\}$ is utilized to determine the attributes of a set of $k$ Gaussians with the embedded latent feature $\hat{f}_v$.
    The attributes, including color, opacity, semantic probability, etc. of the 3D Gaussians, are then derived by decoding the anchor features $\hat{f}_v$ and the corresponding position between the anchor point and the camera center using respective lightweight MLP decoders (\emph{e.g.} opacity decoder $\mathbf{F}_{\alpha}$, semantic decoder $\mathbf{F}_{\text{sem}}$). 
    For example, the opacity and the semantic probability of the relative Gaussians of an anchor point can be obtained by:
    \begin{equation}\label{eq:decoder}
        \{\alpha_1,...,\alpha_k\}=\mathbf{F}_{\alpha}(\hat{f_v}, \delta_{vc}, \textbf{b}_{vc})
    \end{equation}
    \begin{equation}
        \{\hat{y}_1,...,\hat{y}_k\}=\mathbf{F}_{\text{sem}}(\hat{f_v}, \delta_{vc}, \textbf{b}_{vc})
    \end{equation}
    where $\alpha$ denotes the opacity, $\hat{y}$ denotes the semantic probability, $\delta_{vc}, \textbf{b}_{vc}$ respectively denotes the distance and direction vector between the anchor point and the camera center.
    The semantic and instance probabilities of the Gaussians are rasterized and rendered in the same way as Eq.~(\ref{eq:render equation}) to obtain the rendered semantic and instance masks.

\subsection{Robust Semantic Anchor Point Initialization}~\label{sec:robust semantic initialization}

    In order to mitigate the negative effects of the noise of the semantic masks, we propose to robustly initialize the semantic anchor point via filtering out the inconsistent parts of the generated masks across different viewpoints.
    Firstly, we generate consistency-enhanced 2D masks by comparing semantic masks between their adjacent viewpoints, locally filtering out the noise.
    Additionally, we project the locally consistent semantic masks into 3D space and smooth them by clustering to generate globally reliable semantic point cloud for initialization and further regularization.
    To supervise newly grown Gaussians, at the initial training stage, we project the semantic point cloud onto the viewpoints to generate consistent semantic masks for supervision.

\noindent \textbf{Consistency-enhanced 2D Mask Generation}\label{sec:consistency-enhanced 2D mask}
    Given datasets with continuous viewpoints, for an arbitrary reference viewpoint, we initially establish a window size $N_w$ to denote the number of neighboring viewpoints $\{\mathcal{P}_i\}_{i=1}^{N_w}$ for comparison. 
    Subsequently, we warp the semantic labels of a neighboring viewpoint onto the reference image plane for comparison.
    For an arbitrary pixel $\mathbf{u} \in \mathcal{P}_i$ with a semantic label $s_\mathbf{u}$ and a depth value $d$, it is projected to 3D with world coordinate $\text{P}_{\text{world}}$:
    \begin{equation}
    \begin{bmatrix}
           \text{P}_{\text{world}} \\
           1
       \end{bmatrix} = {\mathbf{T}^c_w}_{i} \cdot \begin{bmatrix}
            \mathbf{o} + {\mathbf{d}} \cdot d \\
            1
        \end{bmatrix} \label{eq:ProjectionEquation}
    \end{equation}
    where ${\mathbf{T}^c_w}_{i}$ denotes the homogeneous transformation matrix from camera coordinate to world coordinate of the neighbor view $\mathcal{P}_i$, $\mathbf{o}$ and $\mathbf{d}$ respectively denote the camera center and the ray direction of the corresponding pixel. 
    The point $\text{P}_{\text{world}}$ can then be transformed into the camera coordinate of the reference view and projected to the image plane of the reference view to get the corresponding pixel $\hat{\mathbf{u}}$:
    \begin{align}
        \begin{bmatrix}
            \text{P}_{\text{ref}} \\
            1
        \end{bmatrix} = {\mathbf{T}^w_c}_{\text{ref}} \cdot \begin{bmatrix}
            \text{P}_{\text{world}} \\
            1
        \end{bmatrix}  \\
        \text{Z}_{\text{ref}} \cdot \begin{bmatrix}
            \hat{\mathbf{u}} \\
            1
        \end{bmatrix} = \mathbf{K}_{\text{cam}} \cdot \begin{bmatrix}
            \text{P}_{\text{ref}} \\
            1
        \end{bmatrix}
    \end{align}
    where $\text{P}_{\text{ref}}$ denotes the coordinates in reference camera coordinate system, ${\mathbf{T}_c^w}_{\text{ref}}$ denotes the homogeneous transformation matrix from world coordinate to camera coordinate of the reference view, $\text{Z}_{\text{ref}}$ denotes the z-value of $\text{P}_{\text{ref}}$, $\mathbf{K}_{\text{cam}}$ represents the intrinsic matrix of camera. For clarity, we denote the above warping operation as $\hat{\mathbf{u}}=w(\mathbf{u})$.

    Ultimately, we obtain the corresponding pixel set across the neighbor viewpoints $\Psi(\hat{\mathbf{u}})=\{\mathbf{u}|w(\mathbf{u})=\hat{\mathbf{u}}, \mathbf{u}\in \{\mathcal{P}_i\}_{i=1}^{N_w}\}$. 
    The value of the consistency-enhanced mask is obtained by: 
    \begin{equation}
        q(\hat{\mathbf{u}})=\prod_{\mathbf{u}\in\Psi(\hat{\mathbf{u}})}\mathbbm{1}\{s_{\hat{\mathbf{u}}}=s_{\mathbf{u}}\},  
    \end{equation}
    where $s_{\hat{\mathbf{u}}}$ denotes the semantic label of pixel $\hat{\mathbf{u}}$, $\mathbbm{1}$ denotes the indicator function.

\noindent \textbf{Robust Semantic Anchor Point Discovery}
    With the locally consistent semantic masks filtered with the consistency-enhanced masks, we further propose to generate robust semantic point cloud to introduce global smoothness.
    Specifically, we first project the locally consistent semantic labels into 3D space to form a primary semantic point cloud with the same operation in Eq.~(\ref{eq:ProjectionEquation}).
    However, the initial projection yields a substantially excessive number of points, resulting in redundancy and significant demand for memory.
    To alleviate this, we employ K-means~\cite{lloyd1982least} methods to cluster and smooth the primary semantic point cloud, largely reducing the number of the points.
    The semantic label of each cluster point is then obtained by a majority voting mechanism among the nearest corresponding points of the cluster centers.
    Subsequently, the generated semantic point cloud is used for initialization and regularization in \ref{sec:initial and regularize}.

\noindent \textbf{Reliable Anchor Point Regularization}\label{sec:initial and regularize}
    We subsequently utilize the robust semantic point cloud for initialization.
    As described in Sec.~\ref{sec:structured scene representation}, the positions of anchor points are initialized by voxelization.
    However, since the anchor points do not explicitly possess the property of semantic labels, it's not feasible to directly assign the labels from the semantic point cloud to anchor points. 
    Therefore, we introduce a cluster loss to regularize the semantic logits of the Gaussians distributed by the initial anchor points:
    \begin{equation}\label{eq:Cluster Loss}
        \mathcal{L}_{\text{cluster}}=
        -\frac{1}{N_\text{ach}\cdot k}
        \sum_{\mathbf{v} \in \textbf{V}} 
        \sum_{\mathbf{g}_\text{s} \in \mathcal{T}(\mathbf{v})}
        s_{\mathbf{v}}\cdot \text{log}(\hat{y}_{\mathbf{g}_\text{s}})
    \end{equation}
    where $\mathbf{g}_\text{s}$ denotes the 3D Gaussians, $\mathcal{T}(\mathbf{v})$ denotes the set of Gaussians determined by the anchor point $\mathbf{v}$, $s_\mathbf{v}$ denotes the semantic one-hot vector of the initial anchor points, $\hat{y}_{\mathbf{g}_\text{s}}$ denotes the semantic probabilities of the relevant Gaussians of the anchor points derived from the semantic decoder $\mathbf{F}_{\text{sem}}$. 
    Benefit from the structure and growing strategy of Scaffold-GS~\cite{lu2023scaffold}, unlike 3DGS~\cite{kerbl20233d}, the position of the anchor points are fixed, which ensures the cluster loss to regularize the semantic field with the unbiased initial semantic point cloud. 
    
    It's noteworthy that the cluster loss only applies regularization to the initial anchor points, while the newly grown Gaussians remain unregulated.
    Therefore, to optimize the newly grown Gaussians, we additionally supervise with consistent and dependable semantic masks generated by projecting the robust semantic point cloud onto the viewpoints before the self-training stage.

\subsection{Self-training with Consistency-enhanced Segments}\label{sec:self-training strategy}

    Although the above methods effectively filter out most of the inconsistency of the input masks, the voting operation ignores the validity of the segment masks, resulting in low-quality edges of segments and different labels unreasonably existing in a certain object.
    Therefore, we propose to integrate the machine-generated segments with the consistently rendered semantic masks to generate more reliable pseudo labels and supervise in a self-training manner.

    Following the initial training stage, PLGS is able to render view-consistent semantic masks.
    For an arbitrary training view, we firstly decompose the machine-generated mask $S_{\text{m2f}}$ into segment regions $\{R_q\}_{q=1}^{N_s}$ according to the different semantic labels, where $N_s$ denotes the number of semantic labels of the mask.
    In light of the probable incorrect semantic label of $S_{\text{m2f}}$, to reduce the effect of the noise, we decompose each $R_q$ into disjointed regions with Region Growing Algorithm~\cite{adams1994seeded} and combine them together to get $\{R_{m}\}_{m=1}^{N_m}$, where $N_m$ denotes the number of the total disjointed areas.
    We assign each region $R_{m}$ the semantic label $s_{\text{pse}}(R_{m})$ by a majority voting operation among the corresponding semantic labels of the rendered mask: 
    \begin{equation}
        s_{\text{pse}}(R_{m})=\textbf{argmax}_{x\in \mathcal{C}}\sum_{\mathbf{u}\in R_{m}} \mathbbm{1}\{ \textbf{argmax}(\hat{y}_{\mathbf{u}})=x\}
    \end{equation}
    where $x$ denotes a specific semantic class, $\mathcal{C}$ denotes the set of the semantic classes, $\hat{y}_{\mathbf{u}}$ denotes the rendered semantic probabilities of the pixel $\mathbf{u}$.
    The generated pseudo labels are then utilized for supervision after the initial training stage.

\subsection{Consistent Instance Mask Generation}\label{sec:consistent instance mask}
    As shown in Fig.~\ref{fig:noisy-pattern}(b), for instance segmentation, the main aspect of the inconsistency is the different instance labels of the same object.
    To effectively lift instance segmentation masks, Panoptic Lifting~\cite{siddiqui2023panoptic} utilizes the Hungarian Algorithm~\cite{kuhn1955hungarian} to match the rendered instance mask $M_{\text{rend}}$ and the machine-generated mask $M_{\text{m2f}}$ with the IoU between the instance segments as the values of cost matrix.
    The matched instance masks are then utilized for supervision.
    However, this strategy relies on a basic assumption that the rendered instance mask shares the same instance label across all training views, which can not be guaranteed when the change of view is significant.
    Therefore, we refer to the operation of semantic masks to supervise the instance field with consistent masks.
    However, different from the direct projection of semantic labels, the instance labels need to be matched and unified first.
    Consequently, we propose to match the instance masks across views in 3D space.

    \begin{algorithm}[tb]
    \caption{Consistent Instance Matching in 3D Space.}
    \label{alg:consistent_instance_generation}
    \begin{algorithmic}[1] 
    \REQUIRE ~~\\ 
        Instance segments of the input instance mask, $\{M_q\}_{q=1}^{N_l}$;\\
        Depth map of the training view, $\text{D}$;\\
        Threshold value for adding new instance, $\tau_{\text{new}};$\\
        Global instance set $\mathcal{I}_{\text{global}}$;
    \ENSURE ~~\\ 
        Updated global instance set $\mathcal{I}'_{\text{global}}$;
        \STATE Initialize local instance set $\mathcal{I}_{\text{local}}=\phi$;
        \FOR{$M_q \in \{M_q\}_{q=1}^{N_l}$}
            \STATE Project $M_q$ into 3D space with $\text{D}$ to generate instance point set $\mathbf{P}_l^{q_l}$;
            \STATE Generate oriented bounding box $\mathbf{B}_l^{q_l}$ for $\mathbf{P}_l^{q_l}$;
        \ENDFOR
        \STATE Update local instance set $\mathcal{I}_{\text{local}}=\{\mathbf{P}_l^{q_l}, \mathbf{B}_l^{q_l}\}_{{q_l}=1}^{N_l}$;
        \IF{$\mathcal{I}_{\text{global}}=\phi$}
            \STATE Update global instance set $\mathcal{I}'_{\text{global}}=\mathcal{I}_{\text{local}}$;
        \ELSE
            \STATE Calculate cost value $O_{q_l}^{q_g}=-\text{IoU}-\text{IoM}$ between $\{\mathbf{B}_l^{q_l}\}_{{q_l}=1}^{N_l}$ and $\{\mathbf{B}_g^{q_g}\}_{{q_g}=1}^{N_g}$;
            \STATE Match local and global pairs with \textit{Hungarian Algorithm};
            \FOR{$q_l$ in $N_l$}
                \IF{$\min_{q_g}{O_{q_l}^{q_g}}>\tau_{\text{new}}$}
                    \STATE Append $\{\mathbf{P}_l^{q_l}, \mathbf{B}_l^{q_l}\}$ to $\mathcal{I}_{\text{global}}$;
                \ELSE
                    \STATE Update each $\{\mathbf{P}_g^{q_g}, \mathbf{B}_g^{q_g}\}$ with matched $\{\mathbf{P}_l^{q_l}, \mathbf{B}_l^{q_l}\}$;
                \ENDIF
            \ENDFOR
        \ENDIF
    \RETURN $\mathcal{I}'_{\text{global}}$;
    \end{algorithmic}
    \end{algorithm}

    As shown in Alg.~\ref{alg:consistent_instance_generation}, for an arbitrary instance mask $M_{\text{m2f}}$, we firstly project each instance segments $M_q \in \{M_q\}_{q=1}^{N_l}$ into 3D space to generate instance point sets $\{\mathbf{P}_l^{q_l}\}_{{q_l}=1}^{N_l}$, where $N_l$ denotes the number of instance segments of the $M_{\text{m2f}}$.
    For each $\mathbf{P}_l^{q_l}$, we generate the oriented bounding box $\mathbf{B}_l^{q_l}$ for the afterward matching operation, which is assumed to be oriented around z-axis for simplification.

    After the process of a single view, we obtain a local instance set $\mathcal{I}_{\text{local}}=\{\mathbf{P}_l^{q_l}, \mathbf{B}_l^{q_l}\}_{{q_l}=1}^{N_l}$.
    We further propose to match the local instance set with the global instance set $\mathcal{I}_{\text{global}}=\{\mathbf{P}_g^{q_g}, \mathbf{B}_g^{q_g}\}_{{q_g}=1}^{N_g}$, which is initialized with the $\mathcal{I}_{\text{local}}$, where $N_g$ denotes the number of the global instance set.
    It's intuitive to represent the relevance between the elements of the two sets with the IoU metric between the oriented bounding boxes.
    However, in most viewpoints, only parts of certain objects are visible, which results in low IoU values for the same object from different viewpoints.
    Therefore, we additionally introduce IoM metric to evaluate whether the partial bounding box is inside the global bounding box for matching.
    Eventually, we utilize the Hungarian Algorithm~\cite{kuhn1955hungarian} to match the point sets of $\mathcal{I}_{\text{local}}$ and $\mathcal{I}_{\text{global}}$ with the cost value defined as 
    
    \begin{align}
        &\text{IoU}(\mathbf{B}_l^{q_l}, \mathbf{B}_g^{q_g})
        = 
        \frac{\|\text{inter}(\mathbf{B}_l^{q_l}, \mathbf{B}_g^{q_g})\|}{\|\mathbf{B}_l^{q_l}\|+\|\mathbf{B}_g^{q_g}\|} \\
        &\text{IoM}(\mathbf{B}_l^{q_l}, \mathbf{B}_g^{q_g})
        =
        \frac{\|\text{inter}(\mathbf{B}_l^{q_l}, \mathbf{B}_g^{q_g})\|}{\text{min}(\|\mathbf{B}_l^{q_l}\|, \|\mathbf{B}_g^{q_g}\|)} \\
        &O_{q_l}^{q_g}
        =
        -\text{IoU}(\mathbf{B}_l^{q_l}, \mathbf{B}_g^{q_g})-\text{IoM}(\mathbf{B}_l^{q_l}, \mathbf{B}_g^{q_g}) 
    \end{align}where $\text{inter}(*,*)$ denotes the intersection box of the input bounding boxes, $\|*\|$ denotes the volume of the bounding box.
    For each matched local instance set $\{\mathbf{P}_l^{q_l}, \mathbf{B}_l^{q_l}\}$, if the minimal cost value $\min_{q_g}O_{q_l}^{q_g}$ is larger than the predefined threshold $\tau_{\text{new}}$, we regard the set as a new instance and append it to $\mathcal{I}_{\text{global}}$.
    Otherwise, we update each global set with matched local set by adding $\mathbf{P}_l^{q_l}$ to $\mathbf{P}_g^{q_g}$ and updating $\mathbf{B}_g^{q_g}$.

    Eventually, by maintaining a global instance set, we obtain a consistent instance point cloud. 
    With the generated consistent semantic mask described in Sec.~\ref{sec:initial and regularize}, we can similarly generate the consistent instance masks $M_{\text{match}}$ of the $things$ classes of generated consistent semantic masks by projection.

\subsection{Loss Functions}
    \noindent \textbf{Reconstruction Loss}
    Following the reconstruction method of Scaffold-GS~\cite{lu2023scaffold}, give an arbitrary viewpoint at an epoch, a whole image is rendered, and the reconstruction loss is then obtained with L1-distance, a D-SSIM term~\cite{wang2004image} between rendered images and ground-truth images and a volume regularization~\cite{lombardi2021mixture, lu2023scaffold}:
    \begin{equation}
        \mathcal{L}_{\text{recon}}=
        \mathcal{L}_{\text{L1}}
        +\lambda_{\text{ssim}}\mathcal{L}_{\text{D-SSIM}}
        +\lambda_{\text{vol}}\cdot \sum_{i=1}^{N_{\text{ng}}}\text{Prod}(\text{diag}(\mathbf{S}_i))
    \end{equation}
    where $N_{\text{ng}}$ denotes the number of the neural Gaussians, $\text{Prod}(\cdot)$ is the product of the values of a vector, $\text{diag}(\cdot)$ denotes the diagonal elements of a matrix, $\mathbf{S}$ is the scaling matrix as described in Sec.~\ref{sec:preliminary}.
    It's noteworthy that only the gradients of the reconstruction loss is back-propagated to the opacity, which implies that we restrict the supervision of the geometry field solely to the RGB image input.
    
    \noindent \textbf{Semantic Loss}
    We supervise the rendered semantic masks in two stages. At the initial stage, we utilize semantic masks generated from the projection of the robust semantic point cloud. At the self-training stage, we generate pseudo-labels for supervision. Besides, as the robust semantic regularization, the $\mathcal{L}_{\text{cluster}}$ is calculated at all training stages. For an arbitrary viewpoint $\mathcal{P}$, the semantic loss is computed by:
    \begin{equation}
        \mathcal{L}_{\text{sem}}=
        -\frac{1}{|\mathcal{P}|}\sum_{\mathbf{u} \in \mathcal{P}} s_{\mathbf{u}}^{\text{tar}}\cdot \log (\hat{y}_{\mathbf{u}})
        + \mathcal{L}_{\text{cluster}}
    \end{equation}
    where $s^{\text{tar}}_{\mathbf{u}}$ denotes the target semantic one-hot vector which is derived from projected semantic label in the initial training stage, and the pseudo label in self-training stage.

    \noindent \textbf{Instance Loss}
    As described in Sec.~\ref{sec:consistent instance mask}, we utilize the instance masks $\hat{M}_{\text{m2f}}$ generated by linear assignment and the consistent instance masks $M_{\text{match}}$ generated by matching in 3D space for supervision.
    \begin{equation}
        \mathcal{L}_{\text{ins}}=
        -\frac{1}{|\mathcal{P}|}\sum_{\mathbf{u}\in \mathcal{P}}
        [\hat{\pi}^{\text{m2f}}_{\mathbf{u}} + \pi^{\text{match}}_{\mathbf{u}}]
        \cdot \log(\hat{h}_{\mathbf{u}})
    \end{equation}
    where for each pixel $\mathbf{u}$ of viewpoint $\mathcal{P}$, $\hat{\pi}^{\text{m2f}}_{\mathbf{u}}$ denotes the instance one-hot vector in $\hat{M}_{\text{m2f}}$, $\pi^{\text{match}}_{\mathbf{u}}$ denotes the instance one-hot vector in $M_{\text{match}}$, $\hat{h}_{\mathbf{u}}$ denotes the rendered instance probability.
    
    The final loss is then obtained by:
    \begin{equation}
        \mathcal{L}=\mathcal{L}_{\text{recon}} + \lambda_{\text{sem}}\cdot \mathcal{L}_{\text{sem}} + \lambda_{\text{ins}}\cdot \mathcal{L}_{\text{ins}}
    \end{equation}

\section{Experiments}
\captionsetup[table]{labelsep=period, textfont=normalfont, labelfont=bf}

\begin{table*}[tp]
\begin{center}
\small
\caption{
    \textbf{Quantitative comparison of reconstruction quality.} 
    \textnormal{We outperform most previous methods and exhibit comparable performance compared to state-of-the-art methods.
    Note that since the semantic and RGB reconstruction of \textit{Contrastive Lift} is the same as \textit{Panoptic Lifting}, we only compare the quality of instance reconstruction.
    Besides, since \textit{Mask2Former} is not scene-aware, it can't be evaluated for PQ$^\text{scene}$.
    }
}
\resizebox{0.85\textwidth}{!}{
    \begin{tabular}{c|ccc|ccc|ccc}  
    \toprule
    \multirow{2}{*}{Method} & \multicolumn{3}{c|}{HyperSim~\cite{roberts2021hypersim}} & \multicolumn{3}{c|}{Replica~\cite{straub2019replica}} & \multicolumn{3}{c}{ScanNet~\cite{dai2017scannet}}  \\ 
    & mIoU$\uparrow$ & PQ$^\text{scene}\uparrow$ & PSNR$\uparrow$ &  mIoU$\uparrow$ & PQ$^\text{scene}\uparrow$ & PSNR$\uparrow$  &  mIoU$\uparrow$ & PQ$^\text{scene}\uparrow$ & PSNR$\uparrow$  \\
    \midrule
    
    Mask2Former~\cite{cheng2022masked} & 53.9& -- & -- & 52.4 & -- & -- & 46.7 & --& -- \\
    
    SemanticNeRF~\cite{zhi2021place} & 58.9 & -- & 26.6 & 58.5 & -- & 24.8 & 59.2 & -- & 26.6 \\
    
    DM-NeRF~\cite{wang2022dm} & 57.6 & 51.6 & 28.1 
        & 56.0 & 44.1 & 26.9
        & 49.5 & 41.7 & 27.5\\
    
    PNF~\cite{kundu2022panoptic} & 50.3 & 44.8 & 27.4 
        & 51.5 & 41.1 & 29.8 
        & 53.9 & 48.3 & 26.7\\
    
    PNF+GT Bounding Boxes & 58.7 & 47.6 & 28.1 
        & 54.8 & 52.5 & \cellcolor{light-yellow}31.6 
        & 58.7 & 54.3 & 26.8\\
    
    Panoptic Lifting~\cite{siddiqui2023panoptic} & \cellcolor{light-red}67.8 & \cellcolor{light-yellow}60.1 & 30.1 
    & 67.2 & \cellcolor{light-orange}57.9 & 29.6 
    & \cellcolor{light-orange}65.2 & \cellcolor{light-orange}58.9 & \cellcolor{light-yellow}28.5\\

    Contrastive Lift~\cite{bhalgat2023contrastive} & 
    -- & \cellcolor{light-orange}62.3 & -- &
    -- & \cellcolor{light-red}59.1 & -- &
    -- & \cellcolor{light-red}62.3 & -- \\
    
    \midrule
    3DGS Baseline & 58.5 & 51.2 & 29.7 
        & \cellcolor{light-yellow}68.2 & 55.7 & \cellcolor{light-orange}36.5 
        & 59.6 & 46.2 & 27.9 \\
    
    3DGS + PanopLift
        & 59.8 & 52.8 & 29.5 
        & \cellcolor{light-orange}68.9 & 56.6 & \cellcolor{light-orange}36.5
        & 61.0 & 47.1 & 27.8 \\

    Scaffold-GS Baseline
        & 59.9 & 50.7 & \cellcolor{light-yellow}33.6
        & 66.8 & 53.8 & \cellcolor{light-orange}36.5
        & 62.0 & 48.4 & \cellcolor{light-orange}28.7\\

    Scaffold-GS + PanopLift
        & \cellcolor{light-yellow}63.4 & 54.6 & \cellcolor{light-red}33.8
        & 67.7 & 54.2 & \cellcolor{light-red}36.6
        & \cellcolor{light-yellow}63.5 & 50.9 & \cellcolor{light-orange}28.7\\
    
    \midrule
    PLGS(Ours) & \cellcolor{light-orange}66.2 & \cellcolor{light-red}62.4 & \cellcolor{light-orange}33.7 
            & \cellcolor{light-red}71.2 & \cellcolor{light-yellow}57.8 & \cellcolor{light-red}36.6 
            & \cellcolor{light-red}65.3 & \cellcolor{light-yellow}58.7 & \cellcolor{light-red}28.8\\
    
    \bottomrule
    \end{tabular}
}
\label{tab:quantative_evaluation}
\end{center}
\end{table*}

\begin{figure*}[t]
    \centering
    \includegraphics[width=0.8\linewidth]{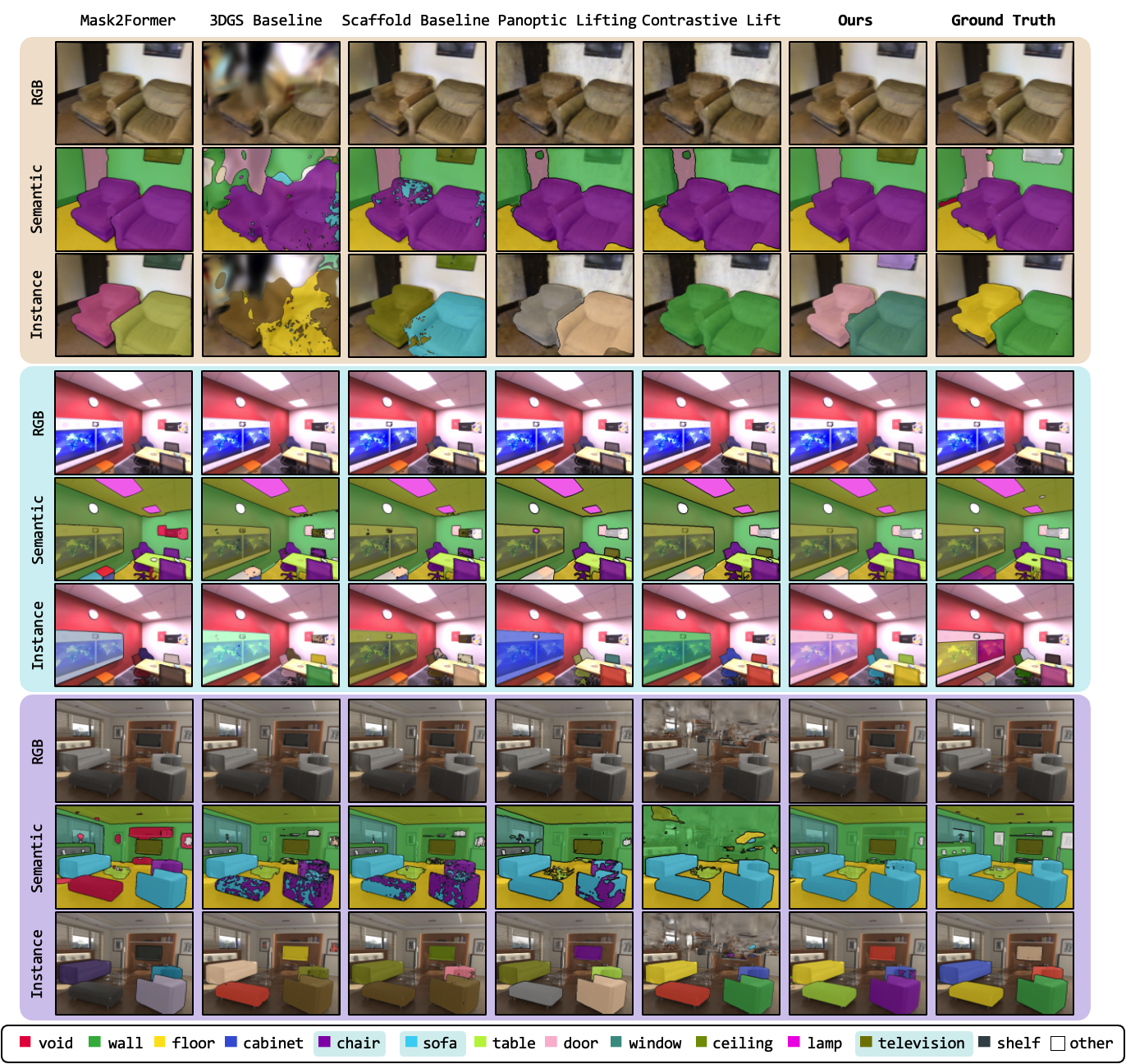}
    \caption{Visualized comparison of novel view synthesis. 
    We demonstrate the RGB, semantic and instance results of our method, previous methods and our baselines on \colorbox{scannet-color}{ScanNet}~\cite{dai2017scannet}, \colorbox{replica-color}{Replica}~\cite{straub2019replica} and \colorbox{hypersim-color}{HyperSim}~\cite{roberts2021hypersim}. 
    The legend labels of \textit{things} classes are highlighted. }
    \label{fig:main_result}
\end{figure*}

\subsection{Experimental Settings}
\subsubsection{Datasets}
    We comprehensively evaluate our method on three public in-place datasets: Hypersim~\cite{roberts2021hypersim}, Replica~\cite{straub2019replica} and ScanNet~\cite{dai2017scannet}. 
    For each of the datasets, we utilize the available ground-truth depth maps and camera poses, while the ground-truth semantic and instance masks are only used for evaluation. 
    For the convenience of comparison of previous works, we keep most of the settings of dataset the same as Panoptic Lifting~\cite{siddiqui2023panoptic}.
    For example, to obtain the machine-generated panoptic masks, we use the 2D panoptic segmentation model Mask2Former~\cite{cheng2022masked}, whose pre-trained weight is trained on COCO~\cite{lin2014microsoft}, to generate the panoptic masks.
    The semantic classes of COCO are then mapped to 21 ScanNet classes.
    The maximum number of instances is set to 25.
    The selected scenes of the datasets are also the same.

\subsubsection{Baselines}\label{sec:baselines}
    In order to demonstrate the feasibility of our method, we conducted a series of baseline experiments.

\noindent \textbf{3DGS/Scaffold-GS Baseline}
    We modified 3DGS~\cite{kerbl20233d} and Scaffold-GS~\cite{lu2023scaffold} for panoptic lifting by embedding semantic and instance label of each Gaussian for 3DGS and adding semantic and instance decoder for Scaffold-GS respectively.
    The baseline methods are initialized with COLMAP~\cite{schonberger2016structure}, and supervised with noisy semantic and instance masks.
    However, since directly training with inconsistent instance masks results in meaningless masks, we adopt the linear assignment method as described in Sec.~\ref{sec:consistent instance mask} for supervision as baseline.

\noindent \textbf{3DGS/Scaffold-GS + Panoptic Lifting}
    To investigate the effect of previous NeRF-based transferable noisy handling methods on 3DGS, we propose to integrate the methods proposed by Panoptic Lifting~\cite{siddiqui2023panoptic} with 3DGS and Scaffold-GS as our additional baseline.
    Specifically, following Panoptic Lifting, we train 3DGS with robust TTA-augmented dataset, and further introduce segment loss proposed in Panoptic Lifting to enhance the quality of semantic segmentation.
    For instance segmentation, we keep the linear-assignment strategy.
    
\subsubsection{Metrics}
    We measure the visual fidelity of the synthesized novel views with peak signal-to-noise ratio (PSNR).
    To evaluate the quality of semantic reconstruction, we utilize the mean intersection over union (mIoU). However, instead of calculating the average mIoU of all test views, we follow Panoptic Lifting~\cite{siddiqui2023panoptic} to treat all the test views as one image by gathering them together to emphasize the 3D consistency of the scene.
    For the quality of instance reconstruction, while conventional panoptic quality (PQ) fails to evaluate the consistency of instance labels of different views, we follow Panoptic Lifting to use scene-level PQ metric $\text{PQ}^{\text{scene}}$, which is capable of evaluate the scene-level consistency of instance labels.
    Moreover, to show that our method is more efficient in both the training and rendering stages, we additionally evaluate the training time and the rendering speed (FPS).

\subsubsection{Implementation Details.}
    In all experiments, the size of RGB image, semantic mask and instance mask is set to $480 \times 640$. 
    As for optional consistency-enhanced 2D mask generation,
    For ScanNet~\cite{dai2017scannet} and Replica~\cite{straub2019replica} whose viewpoints are continuous, we adopt consistency-enhanced 2D mask generation to refine the 2D masks of each view.
    For HyperSim~\cite{roberts2021hypersim} whose viewpoints are sparse, we directly project the semantic labels into 3D space to generate robust semantic point cloud.
    
    The main part of our model is implemented with Pytorch and the panoptic-aware rasterization part of our model is implemented with CUDA. 
    All parameters of our model are trained using Adam~\cite{kingma2014adam}. 
    For weights of losses, we set $\lambda_{\text{ssim}}=\lambda_{\text{sem}}=\lambda_{\text{ins}}=0.2, \lambda_{\text{vol}}=0.001$. 
    The threshold of adding new instance $\tau_{\text{new}}$ is set to -0.1.
    We set the interval of the generation of pseudo labels and the total training iterations as 5k and 30k respectively.
    The initial training stage is conducted before the first generation of the pseudo labels.
    All experiments are conducted on a single NVIDIA RTX 4090. 

\begin{table}[tp]
    \begin{center}
    \caption{Quantitative comparison on time consumption of 3D models. Our method significantly reduces the training time and accelerates the rendering speed.}
    \resizebox{\linewidth}{!}{
    \begin{tabular}{c|cc}
    \toprule
     \makecell{Method}  & Training Time/h$\downarrow$ & Rendering Speed/FPS$\uparrow$ \\
    \midrule

    DM-NeRF~\cite{wang2022dm}  & 22.3 & 0.1 \\
    
    Panoptic Lifting~\cite{siddiqui2023panoptic} & 24.1 & 0.6 \\

    Contrastive Lift~\cite{bhalgat2023contrastive} & 22.6 & 0.7\\

    3DGS Baseline & 1.9 & \textbf{90.1} \\

    Scaffold-GS Baseline & \textbf{1.7} & 21.0 \\
    
    PLGS(Ours)  & 2.0 & 20.8 \\
    
    \bottomrule
    \end{tabular}
    }

    \label{tab:time consumption}
    \end{center}
\end{table}

\subsection{Comparison with Previous State-of-the-art}
    \label{sec:quality comparison}
    \noindent \textbf{Reconstruction and Segmentation Quality}
    In Tab.~\ref{tab:quantative_evaluation}, we compare the reconstruction quality of our method with previous state-of-the-art methods on three benchmarks including HyperSim~\cite{roberts2021hypersim}, Replica~\cite{straub2019replica}, and ScanNet~\cite{dai2017scannet}. 
    For previous NeRF-based methods include SemanticNeRF~\cite{zhi2021place}, DM-NeRF~\cite{wang2022dm}, PNF~\cite{kundu2022panoptic}, Panoptic Lifting~\cite{siddiqui2023panoptic} and Contrastive Lift~\cite{bhalgat2023contrastive}, we directly cite the numbers reported in their papers.
    Since Mask2Former~\cite{cheng2022masked} is not scene-aware and SemanticNeRF only reconstructs semantic field, they can't be evaluated with PQ$_\text{scene}$.
    Besides, since Contrastive Lift shares the same TensoRF~\cite{chen2022tensorf} architecture for RGB reconstruction and MLP architecture for semantic field as Panoptic Lifting, we only report its PQ$_\text{scene}$.
    As shown in Fig.~\ref{fig:main_result}, we further visualize the output of our baseline methods, previous state-of-the-art methods and our methods for comparison.

    As shown in Tab.~\ref{tab:quantative_evaluation}, in terms of the semantic and instance reconstruction quality, our method performs favourably against previous state-of-the-art methods, \emph{e.g.,} Panoptic Lifting and Contrastive Lift. 
    For example, we outperform Panoptic Lifting by 4.0 in mIoU and 7.0 in PSNR, with only a 0.1 decrease in PQ${_\text{scene}}$ and a 1.3 decrease in PQ${_\text{scene}}$ compared to Contrastive Lift on Replica.
    Compared to 3DGS and Scaffold-GS baseline, we outperform by 5.1 in mIoU and 8.6 in PQ$_\text{scene}$ in average, which proves that naively training 3DGS or Scaffold-GS for panoptic lifting fails to output robust panoptic masks as shown in Fig.~\ref{fig:main_result}.
    Compared to 3DGS and Scaffold-GS baseline enhanced with panoptic lifting, we outperform by 3.5 in mIoU and 4.7 in PQ$_\text{scene}$ in average, which proves that our methods are superior to previous NeRF-based noise handling methods.

    \noindent \textbf{Training and Rendering Speed}
    In Tab.~\ref{tab:time consumption}, we exhibit the training time and rendering speed of our methods compared to 3DGS, Scaffold-GS baseline and previous NeRF-based methods of Replica~\cite{straub2019replica}.
    Specifically, we evaluate all the methods on Replica~\cite{straub2019replica} on the same device with a single NVIDIA RTX 4090.
    
    As shown in Tab.~\ref{tab:time consumption}, our method significantly reduces the training time by over 10 times and accelerates the rendering speed by 30 times compared to Panoptic Lifting~\cite{siddiqui2023panoptic}.
    Although Contrastive Lift improves the instance segmentation ability beyond Panoptic Lifting, its speed is still very low.
    PLGS significantly outperforms NeRF-based methods in terms of the training and rendering speed.
    Compared to Scaffold-GS, our method exhibits superior panoptic performance with negligible additional time consumption.
    It can be seen that our main time consumption comes from the operations of Scaffold-GS. 
    The additional costs of noise-handling designs to achieve smooth rendering results are marginal.

\subsection{Ablations}

    To evaluate the effectiveness of our designed methods, we conduct comprehensive ablation experiments on our main designs.

\begin{figure*}[tb]
    \centering
    \includegraphics[width=\linewidth]{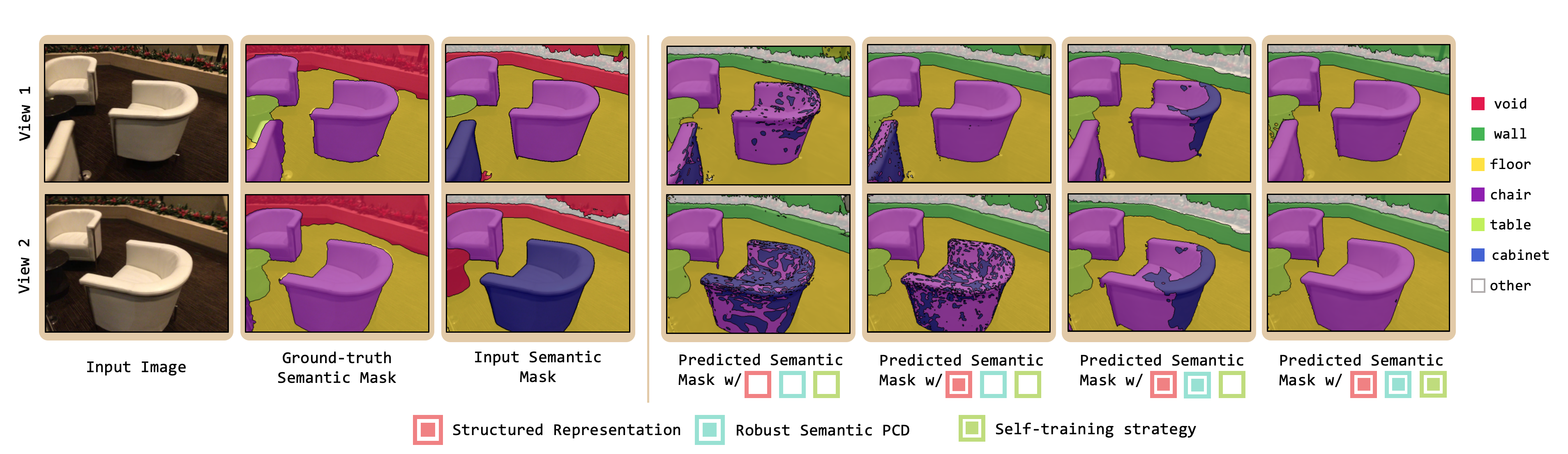}
    \caption{Visualized results of semantic ablation experiments. We demonstrate the results of the semantic ablation experiments by additionally integrating our designed methods to the baseline.}
    \label{fig:semantic_ablation}
\end{figure*}

\begin{figure}[h]
    \centering
    \includegraphics[width=0.85\linewidth]{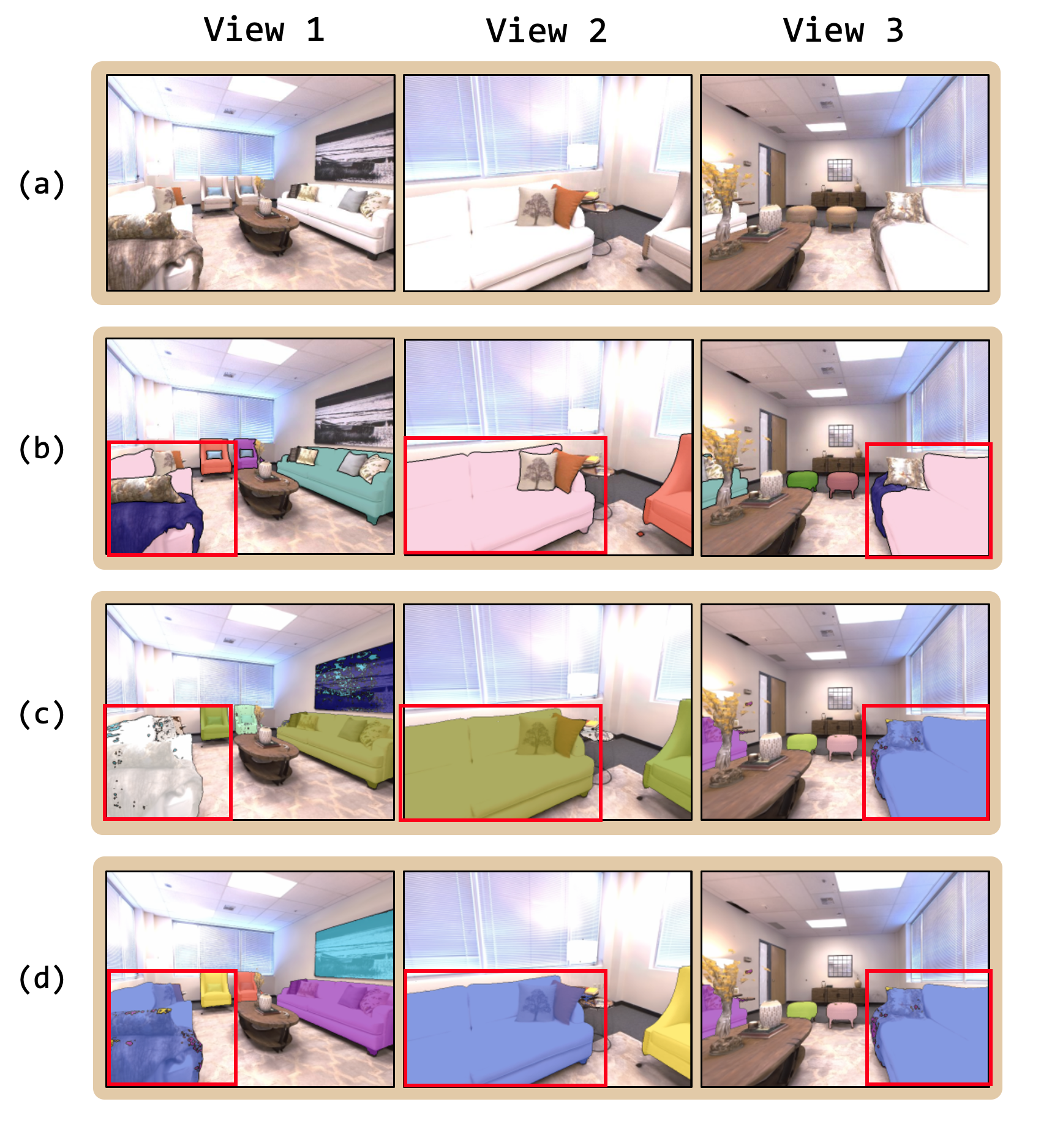}
    \caption{Visualized results of instance ablation experiments. (a) Input RGB images. (b) Ground-truth instance masks. (c) Predicted instance masks without consistent instance guidance. (d) Predicted instance masks with consistent instance guidance. The red boxes emphasize the effectiveness of our designed consistent instance guidance.}
    \label{fig:instance_ablation}
\end{figure}

\subsubsection{Ablations of Main Designs}\label{sec:ablation main design}

    As illustrated in Tab.~\ref{tab:ablations}, we present the results of a series of ablation experiments by incrementally adding our mainly designed methods on the basis of the 3DGS baseline which is decribed in Sec.~\ref{sec:baselines}. 

\begin{table}[h]
    \begin{center}
    \caption{Ablations of our main design on ScanNet~\cite{dai2017scannet}. We conduct ablation experiments by incrementally adding our designs.}
    \resizebox{\linewidth}{!}{
    \begin{tabular}{cccc|cc}
    \toprule
     \makecell{Structured \\ Representation} & \makecell{Robust Semantic\\Initialization}  & \makecell{Self-training\\ Strategy} & \makecell{Consistent\\Instance Guidance}  & mIoU$\uparrow$ & PQ$^\text{scene}\uparrow$ \\
    \midrule
    \XSolidBrush & \XSolidBrush & \XSolidBrush & \XSolidBrush & 59.6 & 46.2 \\
    
    \Checkmark & \XSolidBrush & \XSolidBrush & \XSolidBrush & 62.0 & 48.4 \\
    
    \Checkmark & \Checkmark & \XSolidBrush & \XSolidBrush & 63.1 & 50.6 \\
    
    \Checkmark & \Checkmark & \Checkmark & \XSolidBrush & \textbf{65.6} & 53.4 \\
    
    \Checkmark & \Checkmark & \Checkmark & \Checkmark & 65.3 & \textbf{58.7} \\
    
    \bottomrule
    \end{tabular}
    }
    
    \label{tab:ablations}
    \end{center}
\end{table}

\noindent \textbf{Effect of Structured Representation}
    We replace the vanilla 3DGS structure of baseline with the structured design as described in Sec.~\ref{sec:structured scene representation}.
    As shown in Fig.~\ref{fig:semantic_ablation}, col 5 and Tab.~\ref{tab:ablations}, row 2, there is a 2.4 improvement in mIoU, a 2.2 increase in PQ$_\text{scene}$ and a slight reduction in noise, which proves that the proposed structure is effective but insufficient.

\noindent \textbf{Effect of Robust Semantic Initialization}
    Based on structured panoptic-aware representation, we initialize and supervise the model with robust semantic point cloud and generated consistent semantic masks as described in Sec.~\ref{sec:robust semantic initialization}.
    As depicted in Fig.~\ref{fig:semantic_ablation}, col 6 and Tab.~\ref{tab:ablations}, row 3, with the utilization of robust semantic point cloud, the point noise of the rendered semantic masks is significantly reduced with improvement of a 1.1 increase in mIoU and a 2.2 increase in PQ$_{\text{scene}}$.

\noindent \textbf{Effect of Self-training Strategy}
    Besides the initial training strategy, we introduce a self-training strategy after the initial stage as described in Sec.~\ref{sec:self-training strategy}.
    As shown in Fig.~\ref{fig:semantic_ablation}, col 7 and Tab.~\ref{tab:ablations}, row 4, with the self-training strategy, the rendered semantic masks are not only view-consistent, but also consistent in each segment.
    However, while there is a significant 2.5 increase in mIoU, the PQ$_\text{scene}$ is still low, which proves that for instance reconstruction, it's not enough to achieve high quality merely relying on linear-assignment method.

\noindent \textbf{Effect of Consistent Instance Guidance}
    Based on the linear-assignment methods, we additionally supervise with consistent instance masks as described in Sec.~\ref{sec:consistent instance mask}.
    As exhibited in Fig.~\ref{fig:instance_ablation}(c), although linear assignment strategy achieves consistent instance reconstruction within local viewpoints, it fails to reconstruct consistent instance masks with significant change of viewpoints.
    As shown in Fig.~\ref{fig:instance_ablation}(d) and Tab.~\ref{tab:ablations}, row 5, with consistent instance guidance as constrain, our method produces consistent instance masks within all viewpoints, achieving much better instance reconstruction results with a 5.3 increase in PQ$_\text{scene}$.

\begin{figure}
    \centering
    \includegraphics[width=0.8\linewidth]{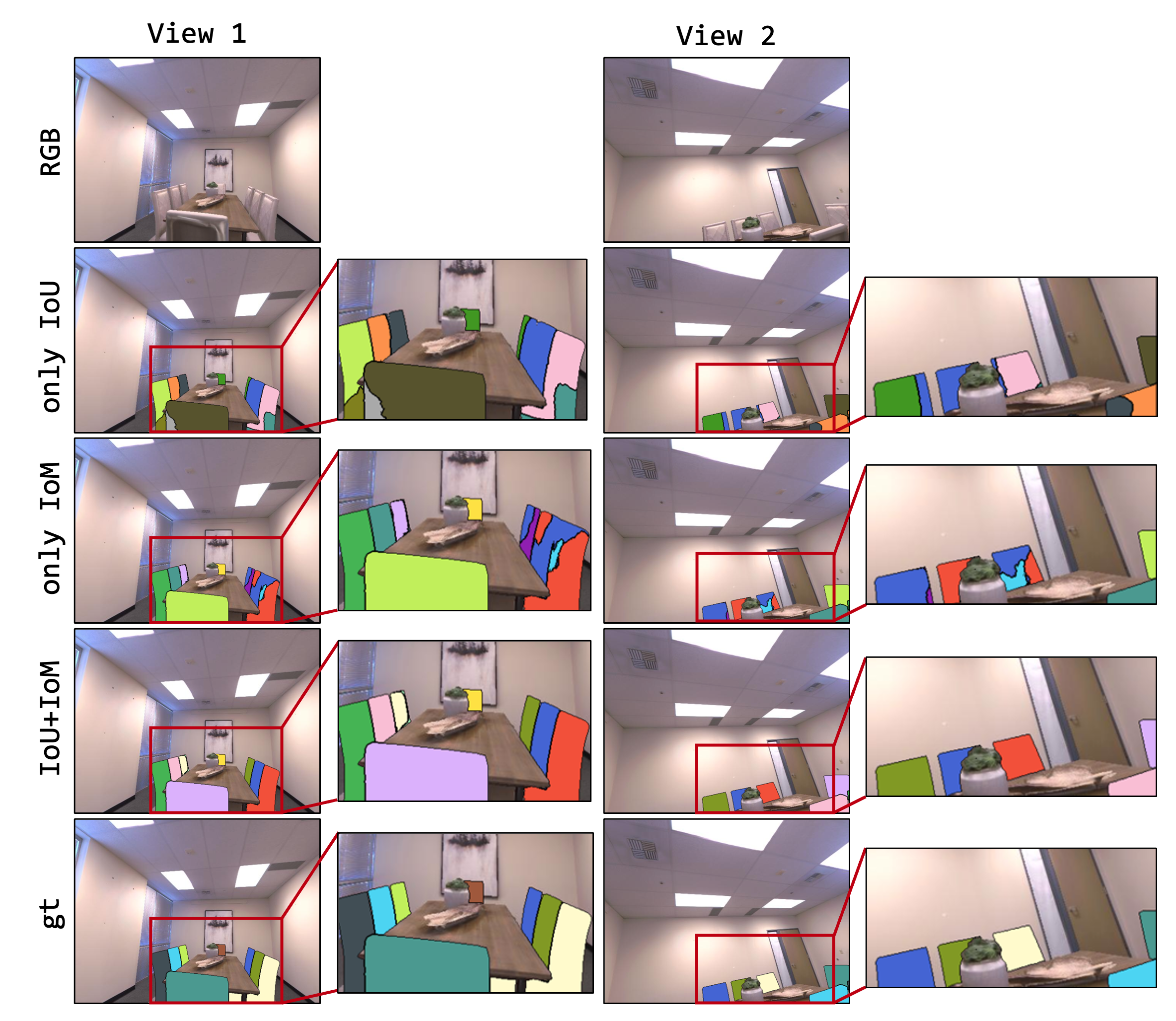}
    \caption{Visualized results of instance masks generated with different design of cost matrix. For clarity, we represent the background of the instance masks with RGB image.}
    \label{fig:cost matrix ablation}
\end{figure}

\subsubsection{Ablations of Consistent Instance Guidance}
    For our proposed instance methods, we validate the efficacy of the design of the cost matrix, and discuss the design of instance loss in the supplementary material.

\begin{table}[h]
    \begin{center}
    \caption{Ablations of cost matrix design on Replica~\cite{straub2019replica}. The reconstruction fails due to the exceeding number of instances when the cost matrix is IoU.}
    \resizebox{0.8\linewidth}{!}{
    \begin{tabular}{c|cc}
    \toprule
     \makecell{Designs of Cost Matrix} & mIoU$\uparrow$ & PQ$^\text{scene}\uparrow$ \\
    \midrule
    IoU & failed & failed \\
    
    IoM & 71.0 & 55.0 \\
    
    IoU+IoM & \textbf{71.2} & \textbf{57.8} \\
    
    \bottomrule
    \end{tabular}
    }
    
    \label{tab:cost matrix ablation}
    \end{center}
\end{table}

\noindent \textbf{Effect of Cost Matrix Design}
    For validation, we compare our design of the cost matrix with only IoU and only IoM.
    Specifically, for each design of cost matrix, we follow the same strategy as described in Sec.~\ref{sec:consistent instance mask}.
    
    As shown in Fig.~\ref{fig:cost matrix ablation} and Tab.~\ref{tab:cost matrix ablation}, using only IoU as the cost value obtains quite low evalutation results because the number of its generated instances is much larger than that of ground-truth instances.
    This is because partially visible instances in most views cannot be accurately matched to the complete instances in $\mathcal{I}_{\text{global}}$ due to the low value of IoU, resulting in them being incorrectly identified as new instances.
    Despite that using only IoM as the cost value can correctly match the partial instances with the whole instances, it leads to lots of mismatches due to the imperfect oriented bounding boxes, which demonstrate a 2.8 decrease in PQ$_{\text{scene}}$ compared to our design.
    However, our method takes both IoU and IoM into consideration, achieving reliable matching results for further supervision.

    We further explore the effect of the hyper-parameter $\tau_{\text{new}}$.
    The excessively high or low values of $\tau_{\text{new}}$ results in misclassification of new instances, thereby affecting the matching performance.
    As shown in Tab.~\ref{tab:valid threshold ablation}, our method is not sensitive to $\tau_{\text{new}}$ and we select -0.1 as the appropriate threshold.

\begin{table}[h]
    \begin{center}
    \caption{Ablations of valid threshold $\tau_{\text{new}}$ on Replica~\cite{straub2019replica}.}
    \resizebox{0.7\linewidth}{!}{
    \begin{tabular}{c|cc}
    \toprule
     \makecell{valid threshold $\tau_{\text{new}}$} & mIoU$\uparrow$ & PQ$^\text{scene}\uparrow$ \\
    \midrule
    -0.05 & 71.1 & 55.7 \\
    
    \textbf{-0.1} & \textbf{71.2} & \textbf{57.8} \\

    -0.2 & 71.1 & 57.5 \\

    -0.4 & 71.0 & 56.2 \\
    
    \bottomrule
    \end{tabular}
    }
    
    \label{tab:valid threshold ablation}
    \end{center}
\end{table}

\subsubsection{Ablations of Utilization of Estimated depths}
To prove the effectiveness of our method in absence of accurate depths, we replace the ground-truth depths with the estimated depths to evaluate our methods.
Specifically, we employ the standard 3DGS~\cite{ye2023gaussian} for preliminary scene reconstruction, utilizing COLMAP's sparse point clouds~\cite{schonberger2016structure} for initialization. 
Furthermore, as officially implemented in 3DGS repository, we follow Hierarchical 3DGS~\cite{hierarchical3DGS} to use the monocular depth estimated via DepthAnythingv2~\cite{depthanythingv2} for additional depth regularization.
Subsequent depth rendering is performed to obtain the final depth maps from the optimized 3D Gaussian representation.

\begin{table}[htbp]
    \begin{center}
    \caption{Quantitative results of our methods with estimated depths on Replica~\cite{straub2019replica}.}
    \resizebox{0.7\linewidth}{!}{
    \begin{tabular}{c|cc}
    \toprule
     \makecell{Method}  & mIoU & PQ$_\text{scene}$ \\
    \midrule
    Scaffold baseline & 66.8 & 53.8 \\
    Ours+Estimated Depth & 71.1 & 57.5 \\
    Ours & \textbf{71.2}  & \textbf{57.8} \\
    \bottomrule
    \end{tabular}
    }
    \label{tab: Results of estimated depth}
    \end{center}
\end{table}

As shown in Tab.~\ref{tab: Results of estimated depth}, our method with estimated depth (``Ours+Estimated Depth'') obtains comparable segmentation performance to our original PLGS with ground-truth depth (71.1 vs. 71.2 in mIoU, 57.5 vs. 57.8 in PQ$_\text{scene}$).
This empirically proves that our approach does not critically depend on high-precision depth measurements, as evidenced by the consistent performance when employing reconstructed depth maps from 3DGS.

\section{Conclusion}
    In this paper, we propose a new framework named PLGS to robustly lift the machine-generated 2D panoptic masks to 3D panoptic masks with remarkable speedup compared to conventional NeRF-based methods.
    Specifically, we build our model based on structured 3D Gaussians and design effective noise reduction strategies to further improve its performance.
    For semantic field, we generate a robust semantic point cloud for initialization and regularization.
    We further design a self-training strategy integrating the input segments and the consistent rendered semantic labels for enhancement.
    For instance field, we utilize oriented bounding boxes to match the inconsistent instance masks in 3D space for consistent instance supervision.
    Experiments show that our method performs favorably against previous NeRF-based panoptic lifting methods with remarkable speedup.
    
\bibliographystyle{IEEEtran}
\bibliography{PLGS}

\end{document}


\date{}
\maketitle

\section{Ablations of Robust Semantic Initialization}

    As shown in Tab.~\ref{tab:semantic_ablations}, we validate the effectiveness of consistency-enhanced masks and 2D consistent semantic masks supervision of the robust semantic initialization section on Replica~\cite{replica}.
    Specifically, we utilize cluster loss as the basic regularization method of the initial semantic point cloud and remove the self-training strategy for all experiments.
    For instance reconstruction, we simply utilize linear-assignment loss as baseline.

\noindent \textbf{Effect of Consistency-enhanced Masks}
    We remove the consistency-enhanced masks and directly project the noisy input semantic masks into 3D space to generate semantic point cloud for validation.
    As shown in Tab.~\ref{tab:semantic_ablations}, row 1-2 and 3-4, the semantic reconstruction quality significantly improves with a 0.8 increase and 4.2 increase in mIoU respectively when the lifted semantic labels are pre-filtered with the consistency-enhanced masks before projection.

\noindent \textbf{Effect of 2D Consistent Semantic Masks}
    We remove the supervision of the generated 2D consistent semantic masks and only regularize the model with cluster loss.
    As shown in Tab.~\ref{tab:semantic_ablations}, row 1,3 and 2,4, with the supervision 2D consistent semantic masks, the newly grown Gaussians can be effectively supervised, which effectively contributes to the quality of the semantic reconstruction with a 0.5 increase and a 3.9 increase in mIoU respectively.

\begin{table}[htb]
    \begin{center}
    \caption{Ablations of the modules of semantic design on Replica~\cite{replica}. We validate each module by systematically removing them one at a time with self-training strategy removed.}
    \resizebox{0.9\linewidth}{!}{
    \begin{tabular}{cc|cc}
    \toprule
     \makecell{Consistency-enhanced\\Masks} & \makecell{2D Consistent\\ Semantic Masks}  & mIoU$\uparrow$ & PQ$^\text{scene}\uparrow$ \\
    \midrule
    \XSolidBrush & \XSolidBrush & 65.5 & 53.0 \\
    
    \Checkmark & \XSolidBrush & 66.3 & 51.8 \\
    
    \XSolidBrush & \Checkmark & 66.0 & 48.9 \\
    
    \Checkmark & \Checkmark & \textbf{70.2} & \textbf{53.1} \\
    
    \bottomrule
    \end{tabular}
    }
    
    \label{tab:semantic_ablations}
    \end{center}
\end{table}

\begin{table}[h]
    \begin{center}
    \caption{Ablations of instance loss design on Replica~\cite{replica}. We demonstrate the results of different design of instance loss.}
    \resizebox{0.9\linewidth}{!}{
    \begin{tabular}{cc|cc}
    \toprule
     \makecell{Linear Assignment} & \makecell{Consistent Instance\\Masks}  & mIoU$\uparrow$ & PQ$^\text{scene}\uparrow$ \\
    \midrule
    \Checkmark & \XSolidBrush & 70.8 & 54.1 \\
    
    \XSolidBrush & \Checkmark & 70.9 & 56.3 \\
    
    \Checkmark & \Checkmark & \textbf{71.2} & \textbf{57.8} \\
    
    \bottomrule
    \end{tabular}
    }
    
    \label{tab:instance loss ablation}
    \end{center}
\end{table}

\section{Ablations of Instance Loss Design}
    In this section, we further discuss the necessity of our instance loss design with specific ablations.
    As shown in Tab.~\ref{tab:instance loss ablation}, we compare our design of instance loss with supervision of linear assignment loss and consistent instance masks individually.
    With only linear assignment loss as supervision, it fails to produce consistent instance masks with significant view changes.
    Although supervision with only consistent instance masks achieves consistent instance reconstruction with better quality, it's limited by the imperfect quality of the oriented bounding boxes and the semantic masks.
    Therefore, our design integrates them together, achieving better quality with a 3.7 increase and a 1.5 increase in PQ$_\text{scene}$ respectively.

\section{Baseline Experiments with gt depths}
To investigate the influence of ground-truth depth on baseline methods, we conduct comprehensive experiments by providing depth information to Panoptic Lifting~\cite{panoptic-lifting}, Contrastive Lift~\cite{contrastlift}, and Scaffold Gaussian baseline with depth-guided rendering loss.
Specifically, for our Scaffold baseline, we utilize ground-truth depths to generate the RGB point cloud for initialization, while for previous NeRF-based works, we minimize the $L_2$ loss between rendered depths and ground-truth depths as additional supervision.

\begin{table}[htbp]
    \begin{center}
    \caption{Quantitative comparison between NeRF-based baselines and 3DGS baseline with depth-supervision on Replica~\cite{replica}. The experiments for Panoptic Lifting and Contrastive Lift are based on their officially released codes.}
    \resizebox{0.8\linewidth}{!}{
    \begin{tabular}{l|cc}
    \toprule
    Method      & mIoU  & PQ$_\text{scene}$ \\
    \midrule
    Panoptic Lifting  & 68.0  & 57.1 \\
    Panoptic Lifting + depth   & 68.2  & 56.9 \\
    Contrastive Lift   & 66.7  & 52.8 \\
    Contrastive Lift + depth   & 67.2  & 53.1 \\
    Scaffold baseline & 66.8 & 53.8 \\
    Scaffold baseline + depth & 67.0 & 53.8 \\
    Ours & \textbf{71.2}  & 57.8 \\
    \bottomrule
    \end{tabular}
    }
    \label{tab: Comparison with depth-supervised baseline}
    \end{center}
\end{table}

As shown in Tab.~\ref{tab: Comparison with depth-supervised baseline}, the depth-supervision has negligible effects on panoptic lifting quality, \emph{e.g.,} introducing depth information to Panoptic Lifting yields 0.2 mIoU gain, but results in a loss of 0.2 PQ$_\text{scene}$. 
The comparisons demonstrate that although depth contains geometric information, it is not trivial to utilize depth to improve the quality of panoptic lifting masks.
We provide an effective way to leverage depth information for robust noise suppression and thus improve the qualit of panoptic lifting masks.

\section{Comparison with Concurrent 3DGS-based works}
We discuss and analyse concurrent 3DGS-based semantic segmentation and object-level segmentation methods: Gaussian grouping~\cite{GaussianGrouping}, Semantic Gaussians~\cite{SemanticGaussians} and Semgauss-SLAM~\cite{Semgauss-SLAM}.

1) \textbf{Comparison with Gaussian-grouping~\cite{GaussianGrouping}.}
Gaussian-grouping is originally designed for the object segmentation lifting tasks and cannot be directly transferred to the panoptic lifting scenario.
Following the authors' instruction and with appropriate modifications, \emph{i.e.,} we utilize DEVA~\cite{DEVA} to refine and match the panoptic masks of Mask2former~\cite{mask2former} and replace the object classifier with semantic classifier and instance classifier,
we conduct experiments to compare our PLGS to Gaussian-grouping.
As demonstrated in Tab.~\ref{tab:comparison with gaussian grouping}, our method remarkably outperforms Gaussian-grouping with 8.1 in mIoU and 11.1 in PQ$_\text{scene}$, demonstrating the superiority of our method in terms of segmentation performance.
In terms of rendering speed, our method is slower than Gaussian-grouping, but much faster than NeRF-based panoptic lifting methods.
Overall, our method strikes a good balance between segmentation performance and rendering speed.

\begin{table}[h]
\caption{Performance comparison between Gaussian-grouping on the Replica~\cite{replica}}
\label{tab:comparison with gaussian grouping}
\centering
\resizebox{\linewidth}{!}{
\begin{tabular}{lcccc}
\toprule
Method & \multicolumn{2}{c}{Panoptic Metrics} & Training Time & Rendering Speeds \\
\cmidrule(lr){2-3} 
 & mIoU (\%) & PQ$_{\text{scene}}$ (\%)  & (h) & (FPS)\\
\midrule
Panoptic Lifting~\cite{panoptic-lifting} & 67.2 & 57.9 & 24.1 & 0.6 \\
Contrastive Lift~\cite{contrastlift} & -- & \textbf{59.1} & 22.6 & 0.7 \\
Gaussian-grouping~\cite{GaussianGrouping} & 63.1 & 46.7 & \textbf{0.5} & \textbf{150}\\
Ours & \textbf{71.2} & 57.8 & 2.0 & 20.8 \\
\bottomrule
\end{tabular}
}
\vspace{0.2cm}
\footnotesize
\end{table}

2) \textbf{Comparison with Semantic Gaussians~\cite{SemanticGaussians}.}
Semantic Gaussians is designed for open-vocabulary semantic lifting task. Specifically, given an off-the-shelf 3D Gaussian Splatting model and the rendered images, it firstly utilizes SAM~\cite{SAM} and the 2D foundation models (\emph{e.g.} CLIP~\cite{CLIP}, OpenSeg~\cite{OpenSeg}) to extract the 2D features.
It then aggregates the features among each mask and projects the pixel-level 2D features to 3D Gaussians with the rendered depths.
It does not adopt any noise-handling techniques, making the performance still constrained by the generalization ability of 2D segmentation model to the viewpoint changes.
To make a comparison with Semantic Gaussians, we only consider the semantic segmentation evalution, 
as shown in Tab.~\ref{tab: Comparison with Semantic Gaussians}. Our method outperforms Semantic Gaussian 3.3 in mIoU on ScanNet~\cite{ScanNet} and achieves comparable rendering speed.

\begin{table}[htbp]
  \centering
  \caption{Semantic Segmentation Results on ScanNet~\cite{ScanNet}}
  \label{tab: Comparison with Semantic Gaussians}
  \begin{tabular}{lcc}
    \toprule
    Method      & \makecell{mIoU\\(\%)} & \makecell{Rendering Speed\\(FPS)} \\
    \midrule
    Semantic Gaussians~\cite{SemanticGaussians}   & 62.0 & \textbf{23.0}   \\
    Panoptic Lifting~\cite{panoptic-lifting}   & 65.2 & 0.6 \\
    Ours & \textbf{65.3} & 20.8 \\
    \bottomrule
  \end{tabular}
\end{table}

3) \textbf{Comparison with Semgauss-SLAM~\cite{Semgauss-SLAM}.}
Semgauss-SLAM~\cite{Semgauss-SLAM} utilizes 3DGS with embedded semantic features to achieve dense semantic SLAM with RGB-D images as input.
However, similar to Semantic Gaussians~\cite{SemanticGaussians}, it directly projects the extracted DINOv2~\cite{DINOv2} features into 3D, ignoring the inconsistency between viewpoints.
Since Semgauss-SLAM is not open-sourced, we didn't compare the results with it.

\section{Experiments with MaskDINO}

\begin{figure}[tbp]
    \centering
    \includegraphics[width=\linewidth]{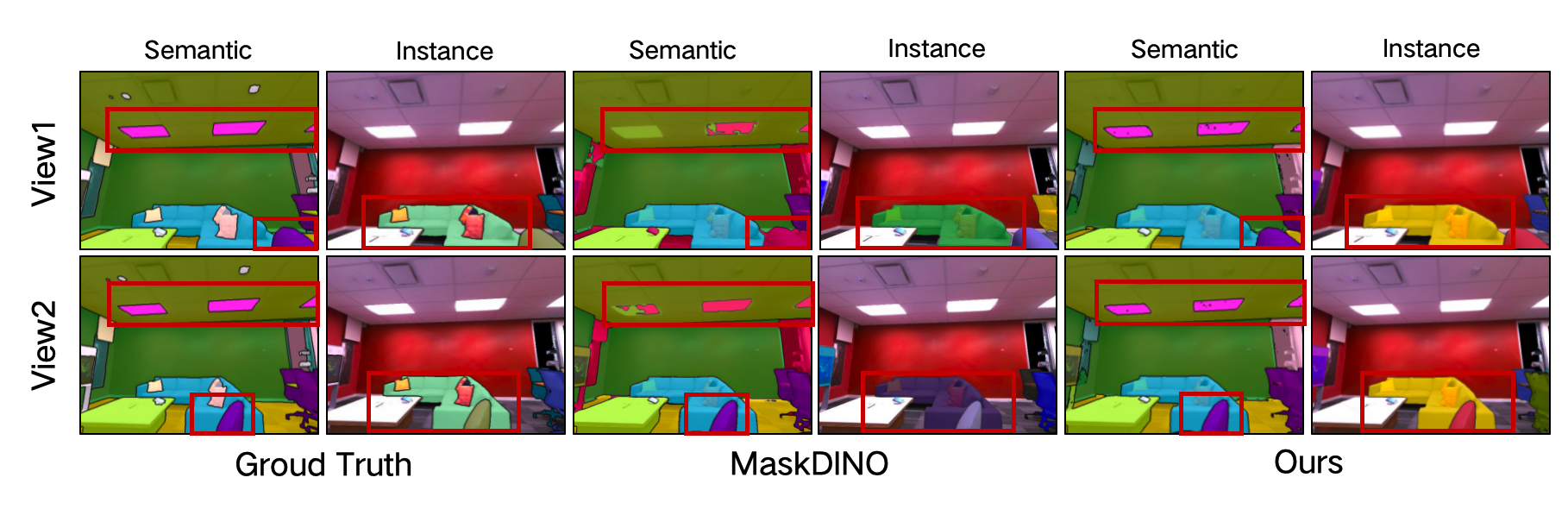}
    \caption{Visualized Comparison with MaskDINO~\cite{MaskDINO}}
    \label{fig: compare with maskdino}
\end{figure}

We conduct experiment with more advanced 2D segmentation model, \emph{i.e.,} MaskDINO~\cite{MaskDINO} to generate the 2D masks as supervision.
As shown in Tab.~\ref{tab: Results on MaskDINO} and Fig.~\ref{fig: compare with maskdino}, although MaskDINO yields better 2D panoptic segmentation masks compared to Mask2Former, there also exists inconsistency across different viewpoints. 
Based on 2D masks generated by MaskDINO, our method also yields remarkable improvement, verifying the effectiveness of our method.

Based on our observations, we believe that due to the lack of 3D prior and imperfect generalization of 2D segmentation model to viewpoint changes, the inconsistency/noise issues in 2D masks always exist, while our solution provides an orthogonal and effective pathway to improve panoptic lifting performance beyond simply developing or adopting more advanced 2D segmentation model.

\begin{table}[htbp]
  \centering
  \caption{Quantitative Results of MaskDINO~\cite{MaskDINO}}
  \label{tab: Results on MaskDINO}
  \begin{tabular}{lccc}
    \toprule
    Method      & mIoU & PQ$_{scene}$ & PQ \\
    \midrule
    MaskDINO~\cite{MaskDINO}   & 53.0 & -- & 49.0  \\
    Ours w MaskDINO & \textbf{65.8} & \textbf{51.0} & \textbf{60.3} \\
    \bottomrule
  \end{tabular}
\end{table}

\section{Comparison with Point Cloud Panoptic Segmentation}
Although our method utilizes an initial point cloud generated using depth information, there are fundamental differences between our approach and point cloud panoptic segmentation models.
First of all, our methods focus on utilizing 3DGS to robustly lift 2D inconsistent machine-generated panoptic masks into 3D, aiming to generate dense rgb images and consistent panoptic masks of arbitrary novel viewpoints.
While point cloud panoptic segmentation methods operate on sparse point clouds, aiming to estimate the labels of each point, which makes it unable to produce dense images and masks.
Therefore, point cloud panoptic segmentation methods can't solve the problem we are concerned about.
Moreover, due to the lack of large-scale 3D point cloud panoptic segmentation data, point cloud panoptic segmentation methods show an inferior ability of generalization compared to 2D panoptic segmentation methods trained on large-scale datasets.

\begin{figure}[htb]
    \centering
    \includegraphics[width=\linewidth]{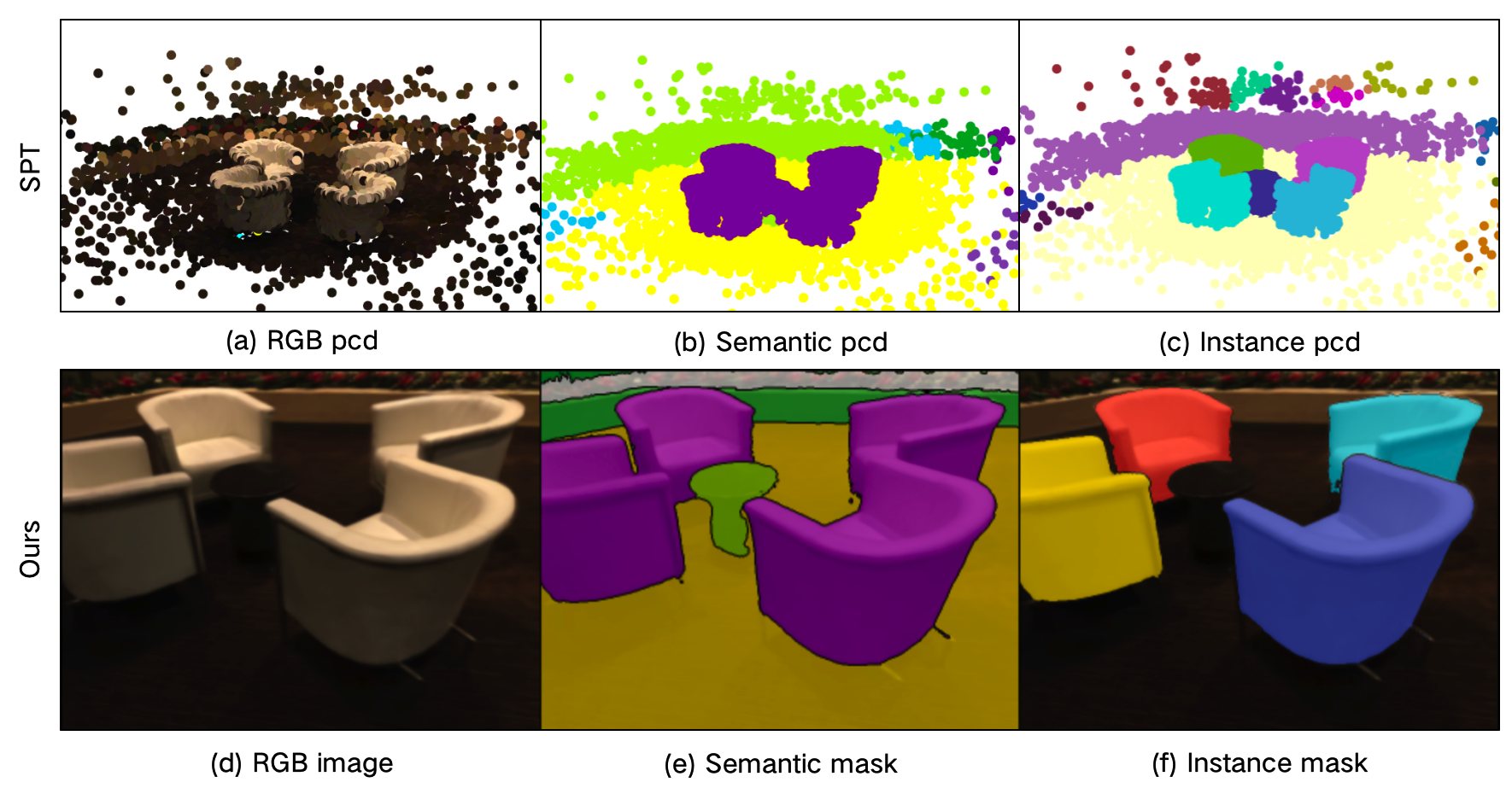}
    \caption{Visualization Results of SPT~\cite{SPT} and our method.}
    \label{fig: pcd vs mask}
\end{figure}

For a more intuitive comparison, we conducte experiments with the recent point cloud panoptic segmentation method SPT~\cite{SPT} to infer the sparse point cloud utilized at the initialization stage of our method.
Specifically, we utilize the model weight pretrained on ScanNet~\cite{ScanNet} and demonstrate the visualized results of a scene of ScanNet, as shown in Fig.~\ref{fig: pcd vs mask}.
Since the point cloud is too sparse to obtain the dense projected 2D masks, we didn't compare the mIoU and PQ$_\text{scene}$ metrics.

\section{More Reconstruction Metrics}
We report the SSIM and LPIPS metrics of our methods and our baseline methods, as shown in Tab.~\ref{tab: reconstruction metric}.
The slight variation of the SSIM, LPIPS metrics empirically proves that our the our methods have slight effect on the reconstruction quality.
It is because we constrain the optimization process, in which only the reconstruction loss is back-propagated to the opacity.

\begin{table}[htbp]
  \centering
  \caption{Reconstruction comparisons on Replica~\cite{replica}.}
  \label{tab: reconstruction metric}
  \begin{tabular}{lcc}
    \toprule
    Method      & SSIM$\uparrow$ & LPIPS$\downarrow$ \\
    \midrule
    3DGS Baseline   & 0.96 & 0.09  \\
    Scaffold-GS Baseline   & 0.96 & 0.07 \\
    Ours & 0.97 & 0.07 \\
    \bottomrule
  \end{tabular}
\end{table}

\printbibliography